%% file: main.tex
\documentclass{article}
\PassOptionsToPackage{numbers, compress}{natbib}
\usepackage[final]{neurips_2020}
\usepackage{times}
\usepackage[hyphens]{url}
\usepackage{float}

\usepackage{graphicx}
\usepackage{microtype}
\usepackage{subfig}
\usepackage{booktabs}
\usepackage{xcolor}
\usepackage{bm}
\usepackage{amsfonts}
\usepackage{mathtools}
\mathtoolsset{showonlyrefs,showmanualtags}
\usepackage{amsthm}
\usepackage{cancel}
\usepackage{thmtools}
\usepackage{authblk}

\declaretheoremstyle[%
  spaceabove=-6pt,%
  spacebelow=6pt,%
  headfont=\normalfont\itshape,%
  postheadspace=1em,%
  qed=\qedsymbol%
]{mystyle} 
\declaretheorem[name={Proof},style=mystyle,unnumbered,
]{prf}

\newtheorem{theorem}{Theorem}
\newtheorem{prop}{Proposition}
\newtheorem{definition}{Definition}
\newtheorem*{remark}{Remark}
\newtheorem{corollary}{Corollary}[theorem]

\usepackage[hidelinks]{hyperref}
\usepackage[flushleft]{threeparttable}

\usepackage{algorithm}
\usepackage{algorithmic}
\usepackage{array}
\usepackage{eqparbox}

\usepackage{etoolbox}
\makeatletter
\patchcmd{\algorithmic}{\addtolength{\ALC@tlm}{\leftmargin} }{\addtolength{\ALC@tlm}{\leftmargin}}{}{}
\makeatother

\newcommand{\mD}{\mathcal D}
\newcommand{\mN}{\mathcal N}
\newcommand{\intd}{\text{ d}}
\newcommand{\KL}{\text{KL}}
\newcommand{\W}{\text{W}}

\newcommand{\M}{\mathcal{M}}

\newcommand{\EPT}{\mathbb{E}}
\newcommand{\erf}{\text{erf}}
\newcommand{\T}{\mathsf{T}}
\newcommand{\ie}{\textit{i.e.}}
\newcommand{\rtext}[1]{\hfill\textcolor{gray}{#1}}

\usepackage{xspace} 

\newcommand{\sref}[2]{\hyperref[#2]{#1~\ref{#2}}}
\newcommand{\eqnref}[1]{\hyperref[#1]{Equation~\eqref{#1}}}

\newcommand{\eat}[1]{}

\newcommand{\revA}[1]{{#1}}
\newcommand{\verify}[1]{{}}

\newcommand{\NOTE}[2]{ }
\newcommand{\TODO}[2]{}
\newcommand{\nb}[1]{}
\newcommand{\mar}[1]{\xspace}

\newcommand\thetitle{Quantile Propagation for \\ Wasserstein-Approximate Gaussian Processes}

\title{\thetitle}
\author[1,2]{Rui Zhang}
\author[1,2]{Christian J. Walder}
\author[2,4]{Edwin V. Bonilla}
\author[2,3]{Marian-Andrei Rizoiu}
\author[1,2]{Lexing Xie}
\affil[1]{The Australian National University}
\affil[2]{CSIRO's Data61, Australia}
\affil[3]{University of Technology Sydney}
\affil[4]{The University of Sydney}
\affil[ ]{\textit{\{rui.zhang, lexing.xie\}@anu.edu.au}, \textit{\{christian.walder, edwin.bonilla\}@data61.csiro.au}}\affil[ ]{\textit{marian-andrei.rizoiu@uts.edu.au}}
\allowdisplaybreaks

\DeclareMathOperator*{\argmax}{argmax}
\DeclareMathOperator*{\argmin}{argmin}

\begin{document}

\maketitle
\setlength{\abovedisplayskip}{5pt}
\setlength{\belowdisplayskip}{5pt}

\input{sections/abstract.tex}
\input{sections/introduction.tex}
\input{sections/relaltedwork.tex}
\input{sections/background.tex}

\input{sections/QP.tex}

\input{sections/Locality.tex}

\input{sections/GOF.tex}
\input{sections/experiments.tex}

\input{sections/conclusion.tex}

\bibliographystyle{apalike}
\bibliography{references}

\input{sections/appendix.tex}
\end{document}

%% file: sections/abstract.tex

\begin{abstract}
Approximate inference techniques are the cornerstone of probabilistic methods based on Gaussian process  priors. Despite this, most work approximately optimizes standard divergence measures such as the Kullback-Leibler (KL) divergence, which lack the basic desiderata for the task at hand, while chiefly offering merely technical convenience.
We develop a new approximate inference method for Gaussian process models which overcomes the technical challenges arising from abandoning these convenient divergences.
Our method---dubbed Quantile Propagation (QP)---is similar to expectation propagation (EP) but minimizes the $L_2$ Wasserstein distance (WD) instead of the KL divergence. The WD exhibits all the required properties of a distance metric, while respecting the geometry of the underlying sample space. 
We show that QP matches quantile functions rather than moments as in EP and has the same mean update but a smaller variance update than EP, thereby alleviating EP's tendency to over-estimate posterior variances.
Crucially, despite the significant complexity of dealing with the WD, QP has the same favorable locality property as EP, and thereby admits an efficient algorithm.
Experiments on classification and Poisson regression show that QP outperforms both EP and variational Bayes.
\end{abstract}

%% file: sections/introduction.tex

\section{Introduction}
Gaussian process (GP) models have attracted \revA{the} attention \revA{of} the machine learning community due to \revA{their} flexibility and \revA{their capacity to} measure uncertainty.
\revA{They} have been widely applied to \revA{learning tasks} such as regression~\citep{10.2113/gsecongeo.58.8.1246}, classification~\citep{williams1998bayesian,hensman2015scalable} and stochastic point process modeling~\citep{moller1998log,Zhang2019}. 
However, exact Bayesian inference for GP models is intractable for all but the Gaussian likelihood function. 
To \revA{address} this issue, various approximate Bayesian inference methods have been  proposed, such as  Markov Chain Monte Carlo~\citep[MCMC, see \textit{e.g.}][]{neal1997monte}, the Laplace approximation~\citep{williams1998bayesian}, variational inference~\citep{jordan1999introduction,opper2009variational} and expectation propagation~\citep{opper2000gaussian,minka2001expectation}. 

The existing approach most relevant to this work is \revA{expectation propagation (EP)}, which approximates each non-Gaussian likelihood term with a Gaussian by iteratively minimizing a set of local forward Kullback-Leibler (KL) divergences. 
As shown by \citet{gelman2017expectation}, EP can scale to very large datasets. 
However, EP \revA{is not} guaranteed \revA{to} converge, and \revA{is known to} over-estimate posterior variances \citep{minka2005divergence,jylanki2011robust,heess2013learning} when approximating a short-tailed distribution. By over-estimation, we mean that the approximate variances are larger than the true variances so that more distribution mass lies in the \textit{ineffective} domain. Interestingly, many popular likelihoods  for GPs results in short-tailed posterior distributions, such as Heaviside and probit likelihoods for GP classification and Laplacian, Student's t and Poisson likelihoods for GP regression.

The tendency to over-estimate posterior variances is an inherent drawback of the forward KL divergence for approximate Bayesian inference. 
More generally, several authors have pointed out that the KL divergence can yield undesirable results such as (but not limited to) over-dispersed or under-dispersed posteriors \citep{dieng2017,li2016renyi,hensman2014tilted}.

As an alternative to the KL, optimal transport metrics---such as the Wasserstein distance \citep[WD,][\textsection 6]{villani2008optimal}---have seen a recent boost of attention. 
The WD is a natural distance between two distributions, and \revA{has been successfully employed in} 
tasks such as image retrieval \citep{rubner2000earth}, text classification \citep{huang2016supervised} and image fusion
\citep{courty2016optimal}. 
Recent work has begun to employ the WD for inference, as in Wasserstein generative adversarial networks
\citep{arjovsky2017wasserstein}, Wasserstein variational inference \citep{ambrogioni2018wasserstein} and Wasserstein  auto-encoders \citep{tolstikhin2017wasserstein}. 
In contrast to the KL divergence, the WD is computationally challenging \citep{cuturi2013sinkhorn}, especially in high dimensions \citep{bonneel2015sliced}, in spite of its intuitive formulation and excellent performance.

\textbf{Contributions}. 
In this work, we develop an efficient approximate Bayesian scheme that minimizes a specific class of WD distances, which we refer to as the $L_2$ WD. Our method overcomes some of the shortcomings of the KL divergence for approximate inference with GP models.
Below we detail \revA{the} three main contributions of this paper.

First, in \autoref{sec:QP}, we develop quantile propagation (QP), an approximate inference algorithm for models with GP priors and factorized likelihoods.
Like EP, QP does not directly minimize global distances between high-dimensional distributions. 
Instead, QP estimates a fully coupled Gaussian posterior by iteratively minimizing \emph{local} divergences between two particular marginal distributions. As these marginals are univariate, QP boils down to an iterative quantile function matching procedure (rather than moment matching as in EP)\,---\,hence we term our 
inference scheme \textit{quantile propagation}. 
We derive the updates for the approximate likelihood terms and show that while the QP mean estimates match those of EP, the variance estimates are lower for QP.

\revA{Second, in} \autoref{sec:locality}  we show that \revA{like EP}, QP satisfies the locality property, \revA{meaning} that it is sufficient to employ \textit{univariate} approximate likelihood terms,  and that the  updates can thereby be performed efficiently using only the marginal distributions.
Consequently, although our method employs a more complex divergence than that of EP ($L_2$ WD vs KL), it has the same computational complexity, after the precomputation of certain (data independent) lookup tables. 

\revA{Finally,} in \autoref{sec:experiment} we employ eight real-world datasets and compare our method to EP  and variational Bayes (VB) on the tasks of binary classification and Poisson regression. 
We find that in terms of predictive accuracy, QP performs similarly to EP but is superior to VB. In terms of predictive uncertainty, however, we find QP superior to both EP and VB, thereby supporting our claim that \revA{QP} alleviates variance over-estimation associated with the KL divergence when approximating  short-tailed distributions \citep{minka2005divergence,jylanki2011robust,heess2013learning}. 

%% file: sections/relaltedwork.tex
%
\section{Related Work}
\label{sec:related_work}

The basis of the EP algorithm for GP models was first proposed by \citet{opper2000gaussian} and then generalized by \citet{minka2001ep,minka2001expectation}.
Power EP \citep{minka2004power,minka2005divergence} is  an extension of EP that exploits the more general $\alpha$-divergence (with $\alpha=1$ corresponding to the forward KL divergence in EP) and \revA{has been} recently \revA{used in conjunction} with  GP pseudo-input approximations \citep{Bui:2017:UFG:3122009.3176848}. 
Although \revA{generally not} guaranteed \revA{to} converge locally or globally, Power EP uses fixed-point iterations for its local updates and has been shown to
perform well in \revA{practice for} GP regression and classification \citep{Bui:2017:UFG:3122009.3176848}. 
\revA{In comparison,} our approach uses the $L_2$ WD, and like EP, \revA{it} yields convex local optimizations \revA{for} GP models with factorized likelihoods. 
This convexity benefits \revA{the} convergence of the local update, and  is retained even with the general $L_p$ ($p \geq 1$) WD as shown in Theorem \ref{thm:convexity}.
Moreover, for the same class of GP models, both EP and our approach have the locality property \citep{seeger2005expectation} and can be unified in the generic message passing framework \citep{minka2005divergence}.

\revA{Without the guarantee of convergence for the explicit global objective function, understanding EP has proven to be a challenging task.}
As a result, 
a number of works \revA{have instead attempted} to directly minimize  divergences between the true and approximate joint posteriors, for divergences such as the KL \citep{jordan1999introduction,dezfouli2015scalable}, R\'enyi \citep{li2016renyi}, $\alpha$ \citep{hernandez2016black} and optimal transport divergences \citep{ambrogioni2018wasserstein}. 
To deal with the infinity issue of the KL (and more generally the R\'enyi and $\alpha$ divergences) which arises from different distribution supports \citep{montavon2016wasserstein,arjovsky2017wasserstein,gulrajani2017improved}, \citet{hensman2014tilted} employ the product of tilted distributions as an approximation.
A number of variants \revA{of EP have also been proposed}, including the convergent double loop algorithm \citep{opper2005expectation}, parallel EP \citep{minka2001family}, distributed EP built on partitioned datasets \citep{xu2014distributed,gelman2017expectation}, averaged EP assuming that all approximate likelihoods contribute similarly \citep{dehaene2018expectation}, and stochastic EP which may be regarded as sequential averaged EP \citep{li2015stochastic}. 

The $L_2$ WD between two Gaussian distributions has a closed form expression \citep{DOWSON1982450}. Detailed research on the Wasserstein geometry of the Gaussian distribution is conducted by \citet{takatsu2011}. Recently, this closed form expression has been applied to robust Kalman filtering \citep{Shafieezadeh-Abadeh:2018:WDR:3327757.3327939} and to the analysis of populations of GPs \citep{NIPS2017_7149}. 
A more general extension to elliptically contoured distributions is provided by \citet{doi:10.1002/mana.19901470121} and used to compute probabilistic word embeddings \citep{Muzellec:2018:GPE:3327546.3327687}. 
A geometric interpretation \revA{for} the $L_2$ WD \revA{between} any distributions \citep{Benamou2000} has already been exploited to develop approximate Bayesian inference schemes \citep{el2012bayesian}. 
Our approach is based on the $L_2$ WD but does not exploit these closed form expressions; instead we obtain computational efficiency by \revA{leveraging} the EP framework and using the quantile function form of the $L_2$ WD for univariate distributions. We believe our work paves the way for further practical approaches to WD-based Bayesian inference.



%% file: sections/background.tex

\section{Prerequisites}
\label{sec:background}

\subsection{Gaussian Process Models}
\label{sec:gpc}
 
Consider a dataset of $N$ samples $\mD = \{\bm x_i, y_i\}_{i=1}^{N}$, where $\bm x_i \in \mathbb{R}^d$ is the input vector and $y_i \in \mathbb R$ is the noisy output. Our goal is to establish the mapping from inputs to outputs via a latent function $f:\mathbb R^d \rightarrow \mathbb R$ which is assigned a GP prior. Specifically, assuming a zero-mean GP prior with covariance function  $k(\bm x, \bm x^\prime; \bm \theta)$, where $\bm \theta$ are the GP hyper-parameters,
we have that $p(\bm f) = \mN(\bm f| \bm 0, K)$, where $\bm f = \{ f_i\}_{i=1}^{N}$, with $f_i \equiv f(\bm x_i)$, is the set of latent function values and
$K$ is the covariance matrix induced by evaluating the covariance function at every pair of inputs. In this work we use the  squared exponential covariance function $k(\bm x, \bm x'; \bm \theta) = \gamma \exp{\big[-\sum_{i=1}^{d}( x_i -  x'_i )^2/(2\alpha_i^2)\big]}$, where $\bm \theta = \{\gamma, \alpha_1,\cdots, \alpha_d \}$.  For simplicity, we will omit \revA{the} conditioning on $\bm \theta$ \revA{in the rest of this paper}. 

Along with the prior, we assume a factorized likelihood $p(\bm y | \bm f) = \prod_{i=1}^N p(y_i | f_i)$ where $\bm y$ is the set of all outputs. Given the above, the posterior $\bm f$ is expressed as:
\begin{align}
p(\bm f | \mD) =p(\mD)^{-1}p(\bm f) \prod_{i=1}^{N} p(y_i | f_i),
\label{eq:GPC_posterior_f}
\end{align} 
where the normalizer $p(\mD)=\int p(\bm f) \prod_{i=1}^{N} p(y_i | f_i) \intd \bm f$ is often analytically intractable. 
Numerous problems take this form: binary classification~\citep{735807}, single-output regression with Gaussian likelihood~\citep{10.2113/gsecongeo.58.8.1246}, Student's-t likelihood~\citep{jylanki2011robust} or Poisson likelihood~\citep{zou2004modified}, and the warped GP \citep{NIPS2003_2481}.

\subsection{Expectation Propagation}
\label{sec:ep}
In this section we review the application of EP to the  GP models described above. EP deals with the analytical intractability by using  Gaussian approximations to the individual non-Gaussian likelihoods:    
\begin{align}
p(y_i|f_i) \approx t_i(f_i)\equiv \widetilde{Z}_i\mN(f_i | \widetilde{\mu}_i, \widetilde{\sigma}_i^2).\label{eq:site_function}   
\end{align}
The function $t_i$ is often called the \textit{site function} and is specified by \revA{the} \textit{site parameters}: the scale $\widetilde{Z}_i$, the mean $\widetilde{\mu}_i$ and the variance $\widetilde{\sigma}_i^2$.
Notably, it is sufficient to use univariate site functions \revA{given that} the local update can be efficiently \revA{computed} using the marginal distribution only \citep{seeger2005expectation}. 
We refer to this as the \emph{locality property}. 
Although \revA{in this work we} employ a more complex $L_2$ WD, our approach retains this property, as we elaborate in \autoref{sec:economic_parameterization_proof}.

Given the site functions, \revA{one can} approximate the intractable posterior distribution $p(\bm f | \mD)$ \revA{using} a Gaussian $q(\bm f)$ as below, where conditioning on $\mD$ is omitted from $q(\bm f)$ for notational convenience:
\begin{align}
q(\bm f)&= q(\mD)^{-1}p(\bm f ) \prod_{i=1}^{N} t_i(f_i)
\equiv \mN(\bm f |\bm \mu, \Sigma ),\label{eq:approximate_posterior}~~\bm \mu = \Sigma \widetilde{\Sigma}^{-1}\widetilde{\bm \mu},~~\Sigma = (K^{-1}+\widetilde{\Sigma}^{-1})^{-1},
\end{align}
where $\widetilde{\bm \mu}$ is the vector of $\widetilde{\mu}_i$, $\widetilde{\Sigma}$ is diagonal with $\widetilde{\Sigma}_{ii}=\widetilde{\sigma}_i^2$;
$\log q(\mD)$ is the log approximate model evidence expressed as below and further employed to optimize GP hyper-parameters: 
{\medmuskip=1mu
	\thinmuskip=1mu
	\thickmuskip=1mu
\begin{align}
    \bm \theta^{\star} = \argmax_{\bm \theta}\, \log\, q(\mD)
    &= \sum_{i=1}^N \log (\widetilde{Z}_i/\sqrt{2\pi}) -\frac{1}{2}\log |K+\widetilde{\Sigma}|-\frac{1}{2}\widetilde{\bm \mu}^{\T} (K+\widetilde{\Sigma})^{-1}\widetilde{\bm \mu}\label{eq:log_pd}.~~~
\end{align}}
The core of EP is to optimize site functions $\{t_i(f_i)\}_{i=1}^N$. 
\revA{Ideally, one would} seek to minimize the global KL divergence between the true and approximate posterior distributions $\KL(p(\bm f|\mD) \| q(\bm f))$, \revA{however this} is intractable. 
Instead, EP is built based on the assumption that the global divergence can be approximated by the local divergence $\KL(\widetilde{q}(\bm f) \| q(\bm f))$, where $\widetilde{q}(\bm f)\propto q^{\setminus i}(\bm f)p(y_i|f_i)$ and $q^{\setminus i}(\bm f) \propto q(\bm f)/t_i(f_i)$ are refered to as the tilted and cavity distributions, respectively. Note that the cavity distribution is Gaussian while the tilted distribution is usually not. The local divergence can be simplified from multi-dimensional to univariate, $\KL(\widetilde{q}(\bm f)\|q(\bm f))=\KL(\widetilde{q}(f_i)\|q(f_i))$ (detailed in \autoref{appx:EP_details}), and $t_i(f_i)$ is optimized by minimizing it.

The minimization is realized by projecting the tilted distribution $\widetilde{q}(f_i)$ onto the Gaussian family,  with the projected Gaussian denoted $\text{proj}_{\KL}(\widetilde{q}(f_i))\equiv\argmin_{\mN} \KL(\widetilde{q}(f_i)\|\mN(f_i))$.  Then the projected Gaussian is used to update $t_i(f_i) \propto \text{proj}_{\KL}(\widetilde{q}(f_i))/q^{\setminus i}(f_i)$.  
The mean and the variance of $\text{proj}_{\KL}(\widetilde{q}(f_i))\equiv \mN(\mu^{\star},\sigma^{\star 2})$ match \revA{the} moments of $\widetilde{q}(f_i)$ and are used to update $t_i(f_i)$'s parameters: 
{\medmuskip=1mu
	\thinmuskip=1mu
	\thickmuskip=1mu
\begin{align}
    \mu^{\star} &= \mu_{\widetilde{q}_i}, \quad \sigma^{\star 2} = \sigma_{\widetilde{q}_i}^2,\label{eq:ep_update}
    \\
    \widetilde{\mu}_i&=\widetilde{\sigma}_i^2 \left(\mu^{\star}(\sigma^{\star})^{ -2} -\mu_{\setminus i}\sigma^{-2}_{\setminus i}\right),~~\widetilde{\sigma}_i^{-2}=(\sigma^{\star})^{-2} - \sigma^{-2}_{\setminus i},~~ \label{eq:ep_update2}
\end{align}}where $\mu_{\widetilde{q}_i}$ and $\sigma_{\widetilde{q}_i}^2$ are the mean and the variance of $\widetilde{q}(f_i)$, and $\mu_{\setminus i}$ and $\sigma_{\setminus i}^2$ are the mean and the variance of $q^{\setminus i}(f_i)$. 
We refer to the projection as the local update. 
\revA{Note} that $\widetilde{Z}$ does not impact the optimization of $q(\bm f)$ or the GP hyper-parameters $\bm \theta$, so we omit the update formula for $\widetilde{Z}$. 
We summarize EP in \sref{algorithm}{alg:ep} (Appendix).
In \autoref{sec:QP} we propose a new approximation approach which is similar to EP but employs the $L_2$ WD rather than the KL divergence. 

\subsection{Wasserstein Distance}
We denote by $\M_+^1(\Omega)$ the set of all probability measures on $\Omega$. We consider probability measures on the $d$-dimensional real space $\mathbb{R}^d$.
The WD between two probability distributions $\xi, \nu \in \M_+^1(\mathbb{R}^d)$ \revA{can be intuitively} defined as the cost of transporting \revA{the} probability mass from one \revA{distribution} to the other. 
We are particularly interested in the subclass of $L_p$ WD, formally  defined as follows.
\begin{definition}[$L_p$ WD]
Consider the set of all probability measures on the product space $\mathbb{R}^d \times \mathbb{R}^d$, whose marginal measures are $\xi$ and $\nu$ respectively, denoted as $U(\xi,\nu)$.
The $L_p$ WD between $\xi$ and $\nu$ is defined as $\W_p^p \left(\xi, \nu \right)\equiv \inf_{\pi \in U(\xi,\nu)} \int_{\mathbb{R}^d\times \mathbb{R}^d} \|\bm x- \bm z\|_p^p \intd \pi(\bm x, \bm z)$ where $p \in [1,\infty)$ and $\|\cdot\|_p$ is the $L_p$ norm.
\end{definition}

Like the KL divergence, the $L_p$ WD it has a minimum of zero, achieved when the distributions are equivalent. \textit{Unlike the KL}, however, it is a proper distance metric, and thereby satisfies the triangle inequality, and has the appealing property of symmetry.

A less fundamental property of the WD which we exploit for computational efficiency is:
\begin{prop}\label{prop:1d}\citep[Remark 2.30]{peyre2019computational}
    The $L_p$ WD between 1-d distribution functions $\xi$ and $\nu \in \mathcal{M}_+^1(\mathbb R)$ equals the $L_p$ distance between the quantile functions,  $\W_p^p(\xi,\nu)=\int_{0}^{1} \left| F_{\xi}^{-1}( y)-F_{\nu}^{-1}(y) \right|^p\intd y$,
    where $F_z:\mathbb R \rightarrow [0,1]$ is the cumulative distribution function (CDF) of $z$, defined as $F_z(x) = \int_{-\infty}^{x} \intd z$, and $F^{-1}_z$ is the pseudoinverse or quantile function, defined as $F_z^{-1}(y) = \min_x \{x\in \mathbb R \cup \{-\infty\}: F_z(x) \geq y \}$.
\end{prop}
Finally, the following translation property of the $L_2$ WD is central to our proof of locality:
\begin{prop}\label{prop:translation}\citep[Remark 2.19]{peyre2019computational}         Consider the $L_2$ WD
    defined for $\xi$ and $\nu \in \mathcal{M}_+^1(\mathbb R^d)$, and let $f_{\bm \tau}(\bm x)=\bm x - \bm \tau$, $\bm \tau \in \mathbb{R}^d$, be a translation operator. If $\xi_{\bm \tau}$ and $\nu_{\bm \tau'}$ denote the probability measures of translated random variables $f_{\bm \tau}(\bm x )$, $\bm x\sim \xi$, and $f_{\bm \tau'}(\bm x )$, $\bm x\sim \nu$, respectively, then $\W_2^2(\xi_{\bm \tau},\nu_{\bm \tau
    '})=\W_2^2(\xi,\nu)-2(\bm\tau-\bm\tau')^{\T}(\bm m_{\xi}-\bm m_{\nu})+\|\bm \tau-\bm \tau'\|_2^2$,
    where $\bm m_{\xi}$ and $\bm m_{\nu}$ are means of $\xi$ and $\nu$ respectively. In particular when $\bm \tau = \bm m_{\xi}$ and $\bm \tau' = \bm m_{\nu}$, $\xi_{\bm \tau}$ and $\nu_{\bm \tau'}$ become zero-mean measures, and $\W_2^2(\xi_{\bm \tau},\nu_{\bm \tau'})=\W_2^2(\xi,\nu)-\|\bm m_{\xi}-\bm m_{\nu}\|_2^2$.
\end{prop}

%% file: sections/QP.tex
\section{Quantile Propagation}
\label{sec:QP}

We now propose our new approximation algorithm which, 
as summarized in \sref{Algorithm}{alg:ep} (Appendix), employs an $L_2$ WD based projection 
rather than the forward KL divergence projection of EP. 
Although QP employs a more complex divergence, it has the same computational complexity as EP, with the following caveat. 
To match the speed of EP, it is necessary to precompute sets of (data independent) lookup tables. Once precomputed, the resulting updates are only a constant factor slower than EP\,---\,a modest price to pay for optimizing a divergence which is challenging \textit{even to evaluate}. Appendix~\ref{sec:lookuptables} provides further details on the precomputation and use of these tables.

As stated in \sref{Proposition}{prop:1d}, minimizing $\W_2^2(\widetilde{q}(f_i), \mN(f_i))$ is equivalent to minimizing the $L_2$ distance between quantile functions of $\widetilde{q}(f_i)$ and $\mN(f_i)$, so we refer to our method as quantile propagation (QP). This section focuses on deriving local updates for the site functions and analyzing their relationships with those of EP. Later in \autoref{sec:locality}, we show the locality property of QP, meaning that the site function $t(f)$ has a univariate parameterization and so the local update can be efficiently performed using marginals only.

\subsection{Convexity of \texorpdfstring{$L_p$}{} Wasserstein Distance}
We first show $\W_p^p(\widetilde{q}(f), \mN(f|\mu,\sigma^2))$ to be strictly convex in $\mu$ and $\sigma$. 
Formally, we have:
\begin{theorem}
Given two probability measures in $\M_+^1(\mathbb{R})$: a Gaussian $\mN(\mu, \sigma^2)$ with  mean $\mu$ and  standard deviation $\sigma>0$, and an arbitrary measure $\widetilde{q}$, 
$\W_p^p(\widetilde{q},\mN)$ is strictly convex in $\mu$ and $\sigma$.
\label{thm:convexity}
\end{theorem}
\begin{prf}
    See \sref{Appendix}{sec:proof_convexity}.
\end{prf}

\subsection{Minimization of \texorpdfstring{$L_2$}{} WD}
\label{sec:min_w22}
An advantage of using the $L_p$ WD with $p=2$, rather than other choices of $p$, is the computational efficiency it admits in the local updates. As we show in this section, optimizing the $L_2$ WD yields neat analytical updates of the optimal $\mu^{\star}$ and $\sigma^{\star}$ that require only univariate integration and the CDF of $\widetilde{q}(f)$. In contrast, other $L_p$ WDs lack convenient analytical expressions. Nonetheless, other $L_p$ WDs may have useful properties, the study of which we leave to future work.

The optimal parameters $\mu^{\star}$ and $\sigma^{\star}$ corresponding to the $L_2$ WD $\W_2^2(\widetilde{q}, \mN(\mu,\sigma^2))$ can be obtained using \sref{Proposition}{prop:1d}. Specifically, we employ the quantile function reformulation of $\W_2^2(\widetilde{q}, \mN(\mu,\sigma^2))$, and zero its derivatives w.r.t. $\mu$ and $\sigma$. The results provided below are derived in \autoref{appx:min_l2_wd}:
\begin{align}
\mu^{\star} = \mu_{\widetilde{q}} ~~;~~
\sigma^{\star} &= \sqrt{2}\int_0^1 F_{\widetilde{q}}^{-1}(y)\erf^{-1}(2y-1) \intd y
=1/\sqrt{2\pi}\int_{-\infty}^{\infty}  e^{-[\erf^{-1}(2F_{\widetilde{q}}(f)-1) ]^2} \intd f.\label{eq:optimal_sigma}
\end{align}
Interestingly, the update for $\mu$ matches that of EP, namely the expectation under $\widetilde{q}$.
However, for the standard deviation we have the difficulty of deriving the CDF $F_{\widetilde{q}}$. If a closed form expression is available, we can apply numerical integration to compute the optimal standard deviation; otherwise, we may use sampling based methods to approximate it. 
\revA{In} our experiments \revA{we employ} the former.

\subsection{Properties of the Variance Update}
\label{sec:varianceproperties}
Given the update equations in the previous section, here we show that  the standard deviation estimate of QP, denoted as $\sigma_{\text{QP}}$, is less or equal to that of EP, denoted as $\sigma_{\text{EP}}$, when projecting the same tilted distribution to the Gaussian space.
\begin{theorem} The variances of the Gaussian approximation to a univariate tilted distribution $\widetilde{q}(f)$ as estimated by QP and EP satisfy $\sigma_{\text{QP}}^2\leq  \sigma^2_{\text{EP}}$. \label{thm:variance}
\end{theorem}
\begin{prf}
See \sref{Appendix}{sec:proof_variance_diff}.
\end{prf}
\begin{corollary}\label{thm:site_post_var}
    The variances of the site functions updated by EP and QP satisfy: $\widetilde{\sigma}^2_{\text{QP}}\leq \widetilde{\sigma}^2_{\text{EP}}$, and the variances of the approximate posterior marginals satisfy $\sigma^2_{q,\text{QP}}\leq \sigma^2_{q,\text{EP}}$. 
\end{corollary}
\begin{prf}
    Since the cavity distribution is unchanged, we can calculate the variance of the site function as per \eqnref{eq:ep_update2} and conclude that the variance of the site function also satisfies $\widetilde{\sigma}_{\text{QP}}^2 \leq \widetilde{\sigma}_{\text{EP}}^2$. Moreover as per the definition of the cavity distribution in \autoref{sec:ep}, the approximate marginal distribution is proportional to the product of the cavity distribution and the site function $q(f_i) \propto q^{\setminus i}(f_i)t(f_i)$, which are two Gaussian distributions. By the product of Gaussians formula (\eqnref{eq:ep_update2}), we know the variance of $q(f_i)$ estimated by EP as $\sigma^2_{q,\text{EP}} = (\widetilde{\sigma}_{\text{EP}}^{-2}+\sigma_{\setminus i}^{-2})^{-1}=\sigma^2_{\text{EP}}$ and similarly $\sigma^2_{q,\text{QP}}=\sigma_{\text{QP}}^2$, where $\sigma_{\text{EP}}^2$ and  $\sigma_{\text{QP}}^2$ are defined in \sref{Theorem}{thm:variance}. Thus, there is $\sigma_{q,\text{QP}}^2 \leq \sigma_{q,\text{EP}}^2$.
\end{prf}
\begin{corollary}\label{thm:pred_var}
    The predictive variances of latent functions at $\bm x_*$ by EP and QP satisfy: $\sigma^2_{\text{QP}}(f(\bm x_*))\leq \sigma^2_{\text{EP}}(f(\bm x_*))$.
\end{corollary} 
\begin{prf}
    The predictive variance of the latent function was analyzed in \citep[Equation (3.61)]{Rasmussen:2005:GPM:1162254}: $\sigma^2 (f_*) = k_*-\bm k_*^{\T}(K+\widetilde{\Sigma})^{-1}\bm k_*$,
where we define $f_* = f(\bm x_*)$ and $k_* = k(\bm x_*, \bm x_*)$, and let $\bm k_* = (k(\bm x_*, \bm x_i))_{i=1}^{N}$ be  the (column) covariance vector  between the test data $\bm x_*$ and the training data $\{\bm x_i\}_{i=1}^{N}$. After updating parameters of the site function $t_i(f_i)$, the predictive variance can be written as (details in \autoref{appx:proof_thm_pred_var}):
\begin{align}
    \sigma_{\text{new}}^2(f_*) = k_*-\bm k_*^{\T} A \bm k_*+ \bm k_*^{\T} \bm s_i \bm s_i^{\T}\bm k_*/[(\widetilde{\sigma}^2_{i,\text{new}}-\widetilde{\sigma}^2_i)^{-1}+A_{ii}],
\end{align}
where  $\widetilde{\sigma}^2_{i,\text{new}}$ is the site variance updated by EP or QP, $A = (K+\widetilde{\Sigma})^{-1}$ and $\bm s_i$ is the $i$'s column of $A$. Since $\widetilde{\sigma}^2_{i,\text{QP}} \leq \widetilde{\sigma}^2_{i,\text{EP}}$, we have $\sigma^2_{\text{QP}}(f_*)\leq \sigma^2_{\text{EP}}(f_*)$.
\end{prf}
\begin{remark}
    We compared variance estimates of EP and QP assuming the same cavity distribution.
    Proving analagous statements for the fixed points of the EP and QP algorithms is more challenging, however, and we leave this to future work, while providing empirical support for these analogous statements in \sref{Figure}{fig:pred_var}. and \sref{Figure}{fig:pred_var_poi}.
\end{remark}

%% file: sections/Locality.tex
\section{Locality Property}
\label{sec:locality}
In this section we detail the central result \revA{on which} our QP algorithm is based \revA{upon}, which we refer to as the   \emph{locality property}. 
\revA{That is}, the optimal site function $t_i$ is defined \revA{only} in terms of the single corresponding latent variable $f_i$, and thereby \revA{and similarly to} EP, \revA{it} admits a simple and efficient sequential update of each individual site approximation.

\subsection{Review: Locality Property of EP}
\label{sec:review_economic_parameterization_ep}
We provide a brief review of the locality property of  EP for  GP models; for more details see \citet{seeger2005expectation}. 
We begin by defining \revA{the} general site function $t_i(\bm f)$ in terms of all of the latent variables, and the cavity and the tilted distributions as $q^{\setminus i} (\bm f) \propto p(\bm f) \prod_{j\neq i} \widetilde{t}_j(\bm f)$ and $\widetilde q(\bm f) \propto q^{\setminus i} (\bm f) p(y_i|f_i)$, respectively. 
To update $t_i(\bm f)$, EP matches a multivariate Gaussian distribution $\mN(\bm f)$ to $\widetilde{q}(\bm f)$ by minimizing the KL divergence $\KL(\widetilde{q} \| \mN)$, which is further rewritten as (see details in Appendix \ref{appx:details_decomposition_KL_proof_EP}): 
\begin{align} \label{eq:KL_locality}
    \KL\big(\widetilde{q} \| \mN&\big)
=\KL\big(\widetilde{q}_i \| \mN_i\big)+\EPT_{\widetilde{q}_i}\Big[\KL\big(q^{\setminus i}_{\setminus i|i} \|\mN_{\setminus i|i}\big)\Big],
\end{align}
where and hereinafter, $\setminus i|i$ denotes the conditional distribution of $\bm f_{\setminus i}$ (taking $f_i$ out of $\bm f$) given $f_i$, namely, $q^{\setminus i}_{\setminus i|i} = q^{\setminus i}(\bm f_{\setminus i}|f_i)$ and $\mN_{\setminus i | i} = \mN(\bm f_{\setminus i} | f_i)$.
Note that $q^{\setminus i}_{\setminus i | i}$ and $\mN_{\setminus i | i}$ in the second term in \eqnref{eq:KL_locality} are both Gaussian, and so setting them equal to one another causes that term to vanish. Furthermore, it is well known that the term $\KL\big(\widetilde{q}_i \| \mN_i\big)$ is minimized w.r.t.~the parameters of $\mathcal N_i$ by matching the first and second moments of $\widetilde{q}_i$ and $\mN_i$. Finally, according to the usual EP logic, we recover the site function $t_i(\bm f)$  by dividing the optimal Gaussian $\mN(\bm f)$ by the cavity $q^{\setminus i}(\bm f)$:
\begin{align}
    t_i(\bm f) &\propto \mN(\bm f)/q^{\setminus i}(\bm f)
 =\cancel{\mN(\bm f_{\setminus i}|f_i)}\mN(f_i)/(\cancel{q^{\setminus i}(\bm f_{\setminus i}|f_i)}q^{\setminus i}(f_i))
=\mN(f_i)/q^{\setminus i}(f_i). \label{eq:show_locality_KL}
\end{align}
Here we can see the optimal site function $t_i(f_i)$ \revA{relies solely} on the local latent variable $f_i$, so it is sufficient to assume a univariate expression for site functions. Besides, the site function can be efficiently updated by using the marginals $\widetilde{q}(f_i)$ and $\mN(f_i)$ only, namely, $t_i(f_i) \propto \big(\min_{\mN_i}\KL(\widetilde{q}_i \| \mN_i))/q^{\setminus i}(f_i)\big)$.

\subsection{Locality Property of QP}
\label{sec:economic_parameterization_proof}
This section proves the locality property of QP, which turns out to be rather more involved to show than is the case for EP. 
\revA{We first prove} the following theorem, and then  follow the same procedure as for EP (\eqnref{eq:show_locality_KL}).
\begin{theorem}
    \label{thm:locality}
Minimization of $\W_2^2 (\widetilde{q}(\bm f), \mN (\bm f))$ w.r.t. $\mN(\bm f)$ results in $q^{\setminus i}(\bm f_{\setminus i}|f_i)=\mN(\bm f_{\setminus i}|f_i)$.
\end{theorem}
\begin{prf}
    See \sref{Appendix}{sec:proof_locality}.
\end{prf}
\begin{theorem}[Locality Property of QP] 
    For GP models with factorized likelihoods, 
    QP requires only univariate site functions, and so yields efficient updates using only marginal distributions.
\end{theorem}
\begin{prf}
     We apply the same steps as in \eqnref{eq:show_locality_KL} for the EP case to QP and we conclude that the site function $t_i(f_i) \propto \mN(f_i)/q^{\setminus i}(f_i)$ relies solely on the local latent variable $f_i$. And as per \eqnref{eq:final_obj} (\sref{Appendix}{sec:proof_locality}), $\mN(f_i)$ is estimated by $\min_{\mN_i} \W_2^2(\widetilde{q}_i,\mN_i)$, so the local update only uses marginals and can perform efficiently.
\end{prf}

\textbf{Benefits \revA{of the} Locality Property.}
The locality property admits an analytically economic form for the site function $t_i(f_i)$, requiring a parameterization that depends on a single latent variable. 
In addition, this also yields a significant reduction in the computational complexity, as only marginals are involved in each local update. In contrast, if QP (or EP) had no such a locality property, estimating the mean and the variance  would involve integrals w.r.t. high-dimensional distributions, with a significantly higher computational cost should closed form expressions be unavailable.

%% file: sections/experiments.tex
\section{Experiments}
\label{sec:experiment}

In this section, we compare the QP, EP and variational Bayes \citep[VB,][]{opper2009variational} algorithms \revA{for} binary classification and Poisson regression. 
The experiments employ eight real world datasets and aim to compare relative accuracy of the three methods, rather than optimizing the absolute performance. The implementations of EP and VB in Python are publicly available \citep{gpy2014}, and our implementation of QP is based on that of EP. Our code is publicly available \footnote{\url{https://github.com/RuiZhang2016/Quantile-Propagation-for-Wasserstein-Approximate-Gaussian-Processes}}. For both EP and QP, we stop local updates, \ie , the inner loop in \sref{Algorithm}{alg:ep} (Appendix), when the root mean squared change in parameters is less than $10^{-6}$. In the outer loop, the GP hyper-parameters are optimized by L-BFGS-B \citep{byrd1995limited} with a maximum of $10^3$ iterations and a relative tolerance of $10^{-9}$ for the function value. VB is also optimized by L-BFGS-B with the same configuration. Parameters shared by the three methods are initialized to be the same.

\subsection{Binary Classification}
\textbf{Benchmark Data.}
We perform binary classification experiments on the five real world datasets employed by \citet{kuss2005assessing}: Ionosphere (IonoS), Wisconsin Breast Cancer, Sonar \citep{Dua:2019}, Leptograpsus Crabs and Pima Indians Diabetes \citep{ripley_1996}. We use two additional UCI datasets as further evidence: Glass and Wine \citep{Dua:2019}. As the Wine dataset has three classes, we conduct binary classification experiments on all pairs of classes. We summarize the dataset size and data dimensions in \autoref{tab:data_results}.

\begin{table*}[ht!]
	\caption{Results on benchmark datasets. The first three columns give dataset names, the number of instances $m$ and the number of features $n$. The table records the test errors (TEs) and the negative test log-likelihoods (NTLLs). The top section is on the benchmark datasets employed by \citet{kuss2005assessing} and the middle section uses additional datasets. The bottom section shows Poisson regression results. * indicates that QP outperforms EP in more than 90\% of experiments \textit{consistently}.}
	\begin{center}
		\setlength{\tabcolsep}{1pt}
		\begin{tabular}{ccccccccc}
			\hline
			 &     &   & \multicolumn{3}{c}{TE ($\times 10^{-2}$)} & \multicolumn{3}{c}{NTLL($\times 10^{-3}$)}
			\\
			\cline{4-9}
			Data &   m   &  n   & EP & QP & VB & EP & QP & VB   
			\\
			\hline
 \texttt{IonoS} & 351 & 34 & $\bm{7.9_{\pm0.5}}$  & $\bm{7.9_{\pm0.5}}$  &$18.9_{\pm6.9}$& $\bm{215.9_{\pm8.4}}$ & $\bm{215.9_{\pm8.5}}$ &$337.4_{\pm70.8}$    \\
 \texttt{Cancer}     & 683 &  9 & $3.2_{\pm0.2}$  & $3.2_{\pm0.2}$ & $\bm{3.1_{\pm0.2}}$ & $\bm{88.2_{\pm3.1}}$       & $\bm{88.2_{\pm3.1}}^*$ &$88.9_{\pm19.1}$ \\
 \texttt{Pima}       & 732 &  7 & $\bm{20.3_{\pm1.0}}$ & $\bm{20.3_{\pm1.0}}$ &$21.9_{\pm0.4}$ & $424.7_{\pm13.0}$      & $\bm{424.0_{\pm13.2}}^*$ &$450.3_{\pm2.6}$\\
 \texttt{Crabs}      & 200 &  7 & $\bm{2.7_{\pm0.5}}$  & $\bm{2.7_{\pm0.5}}$ &$3.7_{\pm0.7}$ &$64.4_{\pm8.2}$      & $\bm{64.3_{\pm8.4}}$ & $164.7_{\pm7.5}$  \\
 \texttt{Sonar}      & 208 & 60 & $\bm{14.0_{\pm1.1}}$ & $\bm{14.0_{\pm1.1}}$ & $25.7_{\pm3.9}$& $306.7_{\pm10.8}$      & $\bm{306.2_{\pm10.9}}^*$ &$693.1_{\pm 0.0}$ \\\hline
 \texttt{Glass} & 214 & 10 & $1.1_{\pm 0.4}$ & $\bm{1.0_{\pm 0.4}}$ &$2.6_{\pm 0.5}$& $29.5_{\pm 5.4}$ & $\bm{29.0_{\pm 5.5}}^*$ & $79.5_{\pm 6.3}$
 \\
 \texttt{Wine1}      & 130 & 13 & $\bm{1.5_{\pm0.5}}$      & $\bm{1.5_{\pm0.5}}$  & $1.7_{\pm0.6}$     & $48.0_{\pm3.4}$   & $\bm{47.4_{\pm3.4}^*}$ & $83.9_{\pm5.2}$      \\
 \texttt{Wine2}      & 107 & 13 & $\bm{0.0_{\pm0.0}}$      & $\bm{0.0_{\pm0.0}}$       & $\bm{0.0_{\pm0.0}}$ & $18.0_{\pm1.2}$  & $\bm{17.8_{\pm1.2}}^*$ &$26.7_{\pm1.9}$     \\
 \texttt{Wine3}      & 119 & 13 & $2.0_{\pm1.0}$      & $2.0_{\pm1.0}$  &  $\bm{1.2_{\pm0.7}}$  & $52.1_{\pm5.6}$ & $\bm{51.8_{\pm5.6}}^*$  &  $69.4_{\pm5.0}$      \\
 \hline
 \texttt{Mining} &  112 & 1 & $\bm{118.6_{\pm 27.0}}$ & $\bm{118.6_{\pm 27.0}}$ & $170.3_{\pm 15.9}$& $1606.8_{\pm 116.3}$& $\bm{1606.5_{\pm 116.3}}$ &$2007.3_{\pm 119.8}$
 \\
 \hline
		\end{tabular}
	\end{center}
	\begin{tablenotes}
		\small
		\item \hskip 1.5cm Note: \texttt{Wine1}: Class 1 vs. 2. \texttt{Wine2}: Class 1 vs. 3. \texttt{Wine3}: Class 2 vs. 3.
	\end{tablenotes}
	\label{tab:data_results}
\end{table*}

\textbf{Prediction.} We predict the test labels using models optimized by EP, QP and VB on the training data.
For a test input $\bm x_*$ with a binary target $y_*$, the approximate predictive distribution is written as: $q(y_*|\bm x_*) = \int_{-\infty}^{\infty} p(y_* | f_*)q(f_*) \intd f_*$
where $f_* = f(\bm x_*)$ is the value of the latent function at $\bm x_*$. We use the probit likelihood for the binary classification task, which admits an analytical expression for the predictive distribution and results in a short-tailed posterior distribution. Correspondingly, the predicted label $\hat{y}_*$ is determined by thresholding the predictive probability at $1/2$. 

\textbf{Performance Evaluation.} To evaluate the performance, we employ two measures: the test error (TE) and the negative test log-likelihood (NTLL). The TE and the NTLL quantify the prediction accuracy and uncertainty, respectively. Specifically, they are defined as $(\sum_{i=1}^{m}|y_{*,i}-\hat{y}_{*,i}|/2)/m$ and $-(\sum_{i=1}^{m}\log q(y_{*,i}|\bm x_{*,i}))/m$, respectively, for a set of test inputs $\{\bm x_{*,i}\}_{i=1}^{m}$, test labels $\{y_{*,i}\}_{i=1}^m$, and the predicted labels $\{\hat{y}_{*,i}\}_{i=1}^{m}$. Lower values indicate better performance for both measures. 
%

\textbf{Experiment Settings.}
In the experiments, we randomly split each dataset into 10 folds,  each time using 1 fold for testing and the other 9 folds for training, with features standardized to zero mean and unit standard deviation. We repeat this  100 times for a random seed ranging $0$ through $99$. As a result, there are a total of 1,000 experiments for each dataset. We report the average and the standard deviation of the above metrics over the 100 rounds. 

\textbf{Results.}
The evaluation results are summarized in \autoref{tab:data_results}. The top section presents the results on the datasets employed by \citet{kuss2005assessing}, whose reported TEs match ours as expected. While QP and EP exhibit similar TEs on these datasets, QP is superior to EP in terms of the NTLL. VB under-performs both EP and QP on all datasets except \texttt{Cancer}. The middle section of \autoref{tab:data_results} shoes the results on additional datasets. The TEs are again similar for EP and QP, while QP has lower NTLLs. Again, VB performs worst among the three methods. To emphasize the difference between NTLLs of EP and QP, we mark with an asterisk those results in which QP outperforms EP in more than 90\% of the experiments. Furthermore, we visualize the predictive variances of QP in comparison with those of EP in \sref{Figure}{fig:pred_var}., which shows that the variances of QP are always less than or equal to those of EP, thereby providing empirical evidence of QP alleviating the over-estimation of predictive variances associated with the EP algorithm.

\subsection{Poisson Regression}

\paragraph{Data and Settings.} We perform a Poisson regression experiment to further evaluate the performance of our method. The experiment employs the coal-mining disaster dataset
\citep{jarrett1979note} which has 190 data points indicating the time of fatal coal mining accidents in the United Kingdom from 1851 to 1962. To generate training and test sequences, we randomly assign every point of the original sequence to either a training or test sequence with  equal probability, and this is repeated 200 times (random seeds $0,\cdots,199$), resulting in 200 pairs of training and test sequences. We use the TE and the NTLL to evaluate the  performance of the model on the test dataset. The NTLL has the same expression as that of the Binary classification experiment, but with a different predictive distribution $q(y_*|\bm x_*)$. The TE is defined slightly differently as $(\sum_{i=1}^{m}|y_{*,i}-\hat{y}_{*,i}|)/m$. To make the rate parameter of the Poisson likelihood non-negative, we use the square link function \citep{flaxman2017poisson,walder2017fast}, and as a result, the likelihood becomes $p(y|f^2)$. We use this link function because it is more mathematically convenient than the exponential function: the EP and QP update formulas, and the predictive distribution $q(y_* | \bm x_*)$ are available in \sref{Appendices}{appx:square_poisson_likelihood} and \ref{appx:square_poisson_pred}, respectively.

\textbf{Results.} The means and the standard deviations of the evaluation results are reported in the last row of \autoref{tab:data_results}. Compared with EP, QP yields lower NTLL, which implies a better fitting performance of QP to the test sequences. We also provide the predictive variances in \sref{Figure}{fig:pred_var_poi}., in the variance of QP is once again seen to be less than or equal to that of EP. This experiment further supports our claim that QP alleviates the problem with EP of over-estimation of the predictive variance. 
Finally, once again we find that both EP and QP outperform VB.
\begin{figure}
    \subfloat[Binary Classification]{\includegraphics[width=0.47\textwidth]{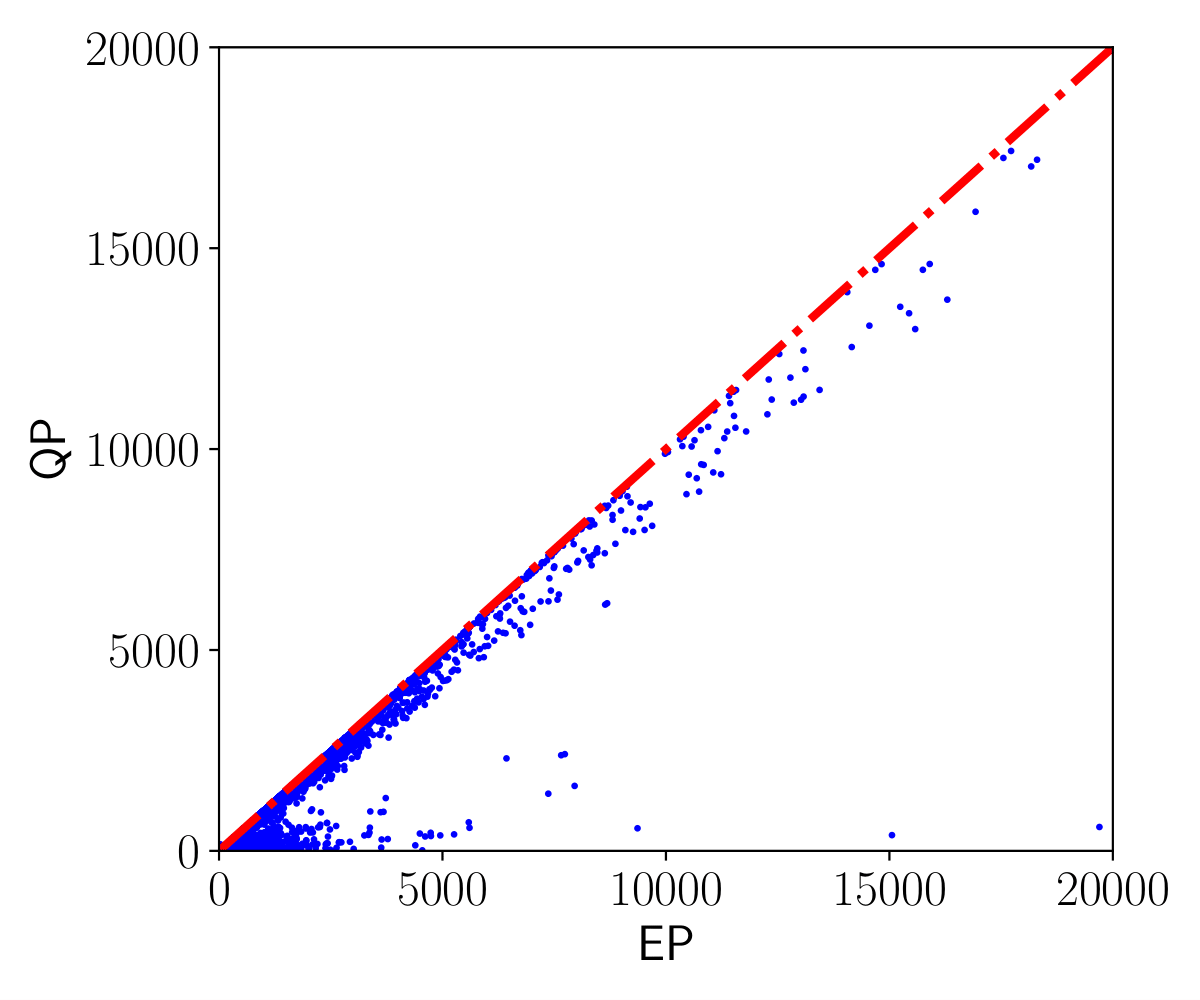}\label{fig:pred_var}}
    \subfloat[Poisson Regression]{\includegraphics[width=0.47\textwidth]{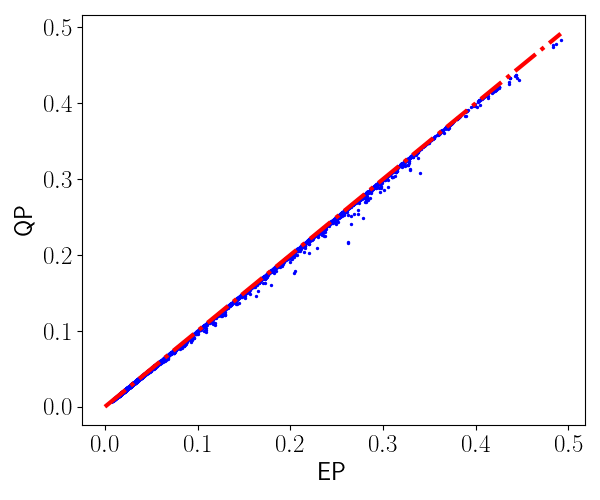}\label{fig:pred_var_poi}}
    \caption{A scatter plot of the predictive variances of latent functions on test data, for EP and QP. The diagonal dash line represents equivalence. We see that the predictive variance of QP is always less than or equal to that of EP.}
\end{figure}

%% file: sections/conclusion.tex

\section{Conclusions}
We have proposed QP as the first efficient $L_2$-WD based approximate Bayesian inference method for Gaussian process models with factorized likelihoods.  Algorithmically, QP is similar to  EP but uses the $L_2$ WD  instead of the forward KL divergence for estimation of the  site functions. When the likelihood factors are approximated by a Gaussian form we show that QP matches quantile functions rather than moments as in EP. Furthermore,  we show that QP  has the same mean update but a smaller variance than that of EP, which in turn alleviates the over-estimation by EP of the posterior variance in practice. Crucially, QP has the same favorable locality property as EP, and thereby admits efficient updates. Our experiments on binary classification and Poisson regression have shown that QP can outperform both EP and variational Bayes. Approximate inference with WD is promising but hard to compute, especially for continuous multivariate distributions. We believe our work paves the way for further practical approaches to WD-based inference.

\paragraph{Limitations and Future Work} Although we have presented properties and advantages of our method, it is still worth pointing out its limitations. First, our method does not provide a methodology for hyper-parameter optimization that is consistent with our proposed WD minimization framework. Instead, for this purpose, we rely on optimization of EP's marginal likelihood. We believe this is one of the reasons for the small performance differences between QP and EP. 

Furthermore, the computational efficiency of our method comes at the price of additional memory requirements and the look-up tables may exhibit instabilities on high-dimensional data. To overcome these limitations,  future work will explore alternatives to hyper-parameter optimization, improvements on numerical computation under the current approach and a variety of WD distances under a similar algorithm framework.

\section*{Broader Impact}
It is likely that the majority of significant technological advancements will eventually lead to both positive and negative societal and ethical outcomes. It is important, however, to consider how and when these outcomes may arise, and whether the net balance is likely to be favourable. After careful consideration, however, we found that the present work is sufficiently general and application independent, as to warrant relatively little specific concern.

\section*{Acknowledgments}
This research was undertaken with the assistance of resources from the National Computational Infrastructure (NCI Australia), an NCRIS enabled capability supported by the Australian Government.
This work is also supported in part by ARC Discovery Project DP180101985 and Facebook Research under the Content Policy Research Initiative grants and conducted in partnership with the Defence Science and Technology Group, through the Next Generation Technologies Program.

%% file: sections/appendix.tex

\newpage
\onecolumn
\appendix
Supplements for
{%
\def\\{\relax\ifhmode\unskip\fi\space\ignorespaces}
\textit{\thetitle}.
}

\section{Minimization of \texorpdfstring{$L_2$}{} WD between Univariate Gaussian and Non-Gaussian Distributions} \label{appx:min_l2_wd}
In this section, we derive the formulas of the optimal $\mu^*$ and $\sigma^*$ for the $L_2$ WD, \ie , Eqn.~\eqref{eq:optimal_sigma}. Recall the optimization problem: we use a univariate Gaussian distribution $\mN(f|\mu,\sigma^2)$ to approximate a univariate non-Gaussian distribution $q(f)$ by minimizing the $L_2$ WD between them:
\begin{equation}
    \min_{\mu,\sigma}\W_2^2(q,\mN)=\min_{\mu,\sigma} \int_0^1 \Big |F_q^{-1}(y)
    -\mu-\sqrt{2} \sigma \erf^{-1}(2y-1)\Big|^2\intd y,
\end{equation}
where $F_q^{-1}$ is the quantile function of the non-Gaussian distribution $q$, namely the pseudoinverse function of the corresponding cumulative distribution function $F_q$ defined in \sref{Proposition}{prop:1d}. 

To solve this problem, we first calculate derivatives about $\mu$ and $\sigma$:
\begin{align}
    \frac{\partial \W_2^2}{\partial \mu}&=-2\int_0^1 F_q^{-1}(y)
    -\mu-\sqrt{2} \sigma \erf^{-1}(2y-1)\intd y,
    \\
    \frac{\partial \W_2^2}{\partial \sigma}&=-2\int_0^1 (F_q^{-1}(y)
    -\mu-\sqrt{2} \sigma \erf^{-1}(2y-1))\sqrt{2}  \erf^{-1}(2y-1)\intd y.
\end{align}
Then, by zeroing derivatives, we obtain the optimal parameters:
\begin{align}
    \mu^* &= \int_0^1 F_q^{-1}(y)-\sqrt{2} \sigma \erf^{-1}(2y-1)\intd y
    \\
    &= \int_{-\infty}^{\infty} x q(x) \intd x-\frac{\sqrt{2}}{2} \sigma \int_{-1}^{1} \erf^{-1}(y)\intd y
    \\
    &= \mu_q-\sqrt{2} \sigma \int_{-\infty}^{\infty}x \mN(x|0,1/2) \intd x
    \\
    &=\mu_q,
    \\
    \sigma^* &= \sqrt{2}\int_0^1 (F_{q}^{-1}(y)-\mu) \erf^{-1}(2y-1) \intd y\Big /\int_0^1 2(\erf^{-1})^2(2y-1) \intd y
    \\
    &= \sqrt{2}\int_0^1 F_q^{-1}(y) \erf^{-1}(2y-1) \intd y\Big /\underbrace{\int_{-\infty}^{\infty} 2 x^2 \mN(x|0,1/2) \intd x}_{=1}
    \\
    &= \sqrt{2}\int_0^1 F_{q}^{-1}(y) \erf^{-1}(2y-1) \intd y
    \\
    & = \sqrt{2}\int_{-\infty}^{\infty} f  \erf^{-1}(2F_q(f)-1) \intd F_q(f)
    \\
    &=-\sqrt{\frac{1}{2\pi}}\int_{-\infty}^{\infty} f \intd\, e^{-[\erf^{-1}(2F_{\widetilde{q}}(f)-1) ]^2}
    \\
    &=0+\sqrt{\frac{1}{2\pi}}\int_{-\infty}^{\infty}  e^{-[\erf^{-1}(2F_{\widetilde{q}}(f)-1) ]^2} \intd f.
    \label{eqn:sigmaintegral}
\end{align}

\section{Minimization of \texorpdfstring{$L_p$}{} WD between Univariate Gaussian and Non-Gaussian Distributions}
\label{appx:min_lp_wd}
In this section, we describe a gradient descent approach to minimizing an $L_p$ WD, for $p \neq 2$, in order to handle cases with no analytical expressions for the optimal parameters. Our goal is to use a univariate Gaussian distribution $\mN(f|\mu,\sigma^2)$ to approximate a univariate non-Gaussian distribution $q(f)$. Specifically, we seek the minimiser in $\mu$ and $\sigma$ of $\W_p^p(q,\mN)$; the derivatives of the objective function about $\mu$ and $\sigma$ are:
\begin{align}
    \partial_{\mu} \W_p^p &= -p\int_{0}^{1} |\varepsilon(y)|^{p-1}\text{sgn}(\varepsilon(y)) \intd y=-p\int_{-\infty}^{\infty} |\eta(x)|^{p-1}\text{sgn}(\eta(x)) q(x)\intd x,
    \\
    \partial_{\sigma} \W_p^p &= -p\int_{0}^{1} |\varepsilon(y)|^{p-1}\text{sgn}(\varepsilon(y))\erf^{-1}(2y-1) \intd y=-p\int_{-\infty}^{\infty} |\eta(x)|^{p-1}\text{sgn}(\eta(x))\erf^{-1}(2F_{q}(x)-1)q(x) \intd x.
\end{align}
where for simplification, we define $\varepsilon(y)=F_{q}^{-1}(y)-\mu-\sqrt{2}\sigma\erf^{-1}(2y-1)$ and $\eta(x)=x-\mu-\sqrt{2}\sigma\erf^{-1}(2F_{q}(x)-1)$, with $F_q$ and $F_q^{-1}$ being the CDF and the quantile function of $q$. 
Note the derivatives have no analytical expressions. However, if the CDF $F_q$ is available, we can use the standard numerical integration routines; otherwise, we resort to Monte Carlo sampling. In the framework of EP or QP, $q(x) \propto q^{\setminus i}(x)p(y_i|x)$ and $q^{\setminus i}$ is Gaussian, so we may draw samples from a Gaussian proposal distribution to obtain a simple Monte Carlo method.


\section{Computations for Different Likelihoods}
Given the likelihood $p(y|f)$ and the cavity distribution $q^{\setminus i}(f) = \mN(f|\mu,\sigma^2)$, a stable way to compute the mean and the variance of the tilted distribution $\widetilde{q}(f) = p(y|f)q^{\setminus i}(f)/Z$ where the normalizer $Z = \int_{-\infty}^{\infty} p(y|f)q^{\setminus i}(f) \intd f$, can be found in the software manual of \citet{Rasmussen:2005:GPM:1162254}. We present the key formulae below, for use in subsequent derivations:
\begin{align}
    \partial_{\mu} Z &= \int_{-\infty}^{\infty} \frac{f-\mu}{\sigma^2} p(y|f)\mN(f|\mu,\sigma^2) \intd f 
    \\
    \frac{\partial_{\mu} Z}{Z}  &= \frac{1}{\sigma^2} \int_{-\infty}^{\infty} f \frac{p(y|f)\mN(f|\mu,\sigma^2) }{Z} \intd f - \frac{\mu}{\sigma^2}\int_{-\infty}^{\infty} \frac{p(y|f)\mN(f|\mu,\sigma^2)}{Z} \intd y
    \\
    \frac{\partial_{\mu} Z}{Z}  &= \frac{1}{\sigma^2} \mu_{\widetilde{q}} - \frac{\mu}{\sigma^2} \\
    \Longrightarrow
    \mu_{\widetilde{q}}&= \frac{\sigma^2 \partial_{\mu} Z}{Z} +\mu = \sigma^2 \partial_{\mu}\log Z+\mu,
    \\
    \partial^2_{\mu} Z &= \int_{-\infty}^{\infty} -\frac{1}{\sigma^2} p(y|f)\mN(f|\mu,\sigma^2)+ \left(\frac{f-\mu}{\sigma^2}\right)^2 p(y|f)\mN(f|\mu,\sigma^2) \intd f 
    \\
    \frac{\partial^2_{\mu} Z}{Z}  &= \int_{-\infty}^{\infty} \left(-\frac{1}{\sigma^2}+\frac{\mu^2}{\sigma^4}+\frac{f^2}{\sigma^4}-\frac{2\mu f}{\sigma^4} \right) \frac{p(y|f)\mN(f|\mu,\sigma^2)}{Z}\intd f 
    \\
    \frac{\partial^2_{\mu} Z}{Z} &= -\frac{1}{\sigma^2}+\frac{\mu^2}{\sigma^4}+\frac{1}{\sigma^4}(\sigma^2_{\widetilde{q}}+\mu_{\widetilde{q}}^2)-\frac{2\mu }{\sigma^4}\mu_{\widetilde{q}} 
    \\
    \frac{\partial^2_{\mu} Z}{Z} &= -\frac{1}{\sigma^2}+\frac{\sigma_{\widetilde{q}}^2}{\sigma^4}+\frac{(\mu-\mu_{\widetilde{q}})^2}{\sigma^4}=-\frac{1}{\sigma^2}+\frac{\sigma_{\widetilde{q}}^2}{\sigma^4}+\left(\frac{\partial_{\mu} Z}{Z}\right)^2
    \\
    \Longrightarrow \sigma^2_{\widetilde{q}} &= \sigma^4 \left[\frac{\partial^2_{\mu} Z}{Z}-\left(\frac{\partial_{\mu} Z}{Z}\right)^2\right]+\sigma^2 = \sigma^4 \partial^2_{\mu} \log Z + \sigma^2.
\end{align}

\subsection{Probit Likelihood for Binary Classification}
For the binary classification with labels $y \in \{-1,1\}$, the PDF of the tilted distribution $\widetilde{q}(f)$ with the probit likelihood is provided by \citet{Rasmussen:2005:GPM:1162254}:
\begin{align}\label{eq:closed_form_pdf_bc}
    \widetilde{q}(f)&= Z^{-1}\Phi(fy)\mN(f| \mu,\sigma^2),~~Z= \Phi (z),~~ z = \frac{\mu}{y\sqrt{1+\sigma^2}},
\end{align}
and the mean estimate also has a closed form expression:
\begin{align}\label{eq:closed_form_mu_bc}
    \mu^\star=\mu_{\widetilde{q}}=\mu+\frac{\sigma^2 \mN(z)}{\Phi(z)y\sqrt{1+\sigma^2}}.
\end{align}
As per \eqnref{eq:optimal_sigma}, the computation of the optimal $\sigma^{\star}$ requires the CDF of $\widetilde{q}$, denoted as $F_{\widetilde{q}}$. For positive $y>0$, the CDF is derived as
\begin{align}
    F_{\widetilde{q},y>0}(x)&= Z^{-1}\int_{-\infty}^{x} \Phi\left(fy\right)\mN\left(f | \mu, \sigma^2\right)\intd f
    \\
    &=\frac{Z^{-1}}{2\pi\sigma y}\int_{-\infty}^{\mu}\int_{-\infty}^{x-\mu}\exp\left(-\frac{1}{2}\begin{bmatrix}w\\f\end{bmatrix}^T \begin{bmatrix}v^{-2}+\sigma^{-2}& v^{-2}\\v^{-2}&v^{-2} \end{bmatrix} \begin{bmatrix}w\\f\end{bmatrix}\right) \intd w \intd f
    \\
    &=Z^{-1}\int_{-\infty}^{k}\int_{-\infty}^{h}\mN\left(\begin{bmatrix}w\\f\end{bmatrix} \bigg |\bm 0,\begin{bmatrix}1& -\rho\\ -\rho &1 \end{bmatrix} \right) \intd w \intd f
    \\
    &\overset{\text{(a)}}{=}Z^{-1}\left[\frac{1}{2}\Phi(h)-T\left(h,\frac{k+\rho h}{h\sqrt{1-\rho^2}}\right )+\frac{1}{2}\Phi(k) -T\left(k,\frac{h+\rho k}{k\sqrt{1-\rho^2}}\right )+\eta \right]
    \\
    k&= \frac{\mu}{\sqrt{\sigma^2+1}},~~ h= \frac{x-\mu}{\sigma},~~ \rho=\frac{1}{\sqrt{1+1/\sigma^2}},~~ x\neq \mu, ~~\mu \neq 0,
\end{align}
where the step (a) is obtained by exploiting the work of \citet{owen1956} and  
$T(\cdot,\cdot)$ is the Owen's T function:
\begin{equation}
    T(h,a) = \frac{1}{2\pi} \int_{0}^{a}\frac{\exp \big[-(1+x^2)h^2/2\big]}{1+x^2}\intd x,
\end{equation}
 and $\eta$ is defined as
\begin{align}
    \eta = \begin{cases}
    0 & \text{$hk > 0$ or ($hk = 0$ and $h + k \geq 0$)}, \\
    -0.5 & \text{otherwise}.
    \end{cases}
\end{align}
Similarly, for $y<0$, the CDF is
\begin{align}
    F_{\widetilde{q},y<0}(x)&=Z^{-1}\left[\frac{1}{2}\Phi(h)+T\left(h,\frac{k+\rho h}{h\sqrt{1-\rho^2}}\right )-\frac{1}{2}\Phi(k) +T\left(k,\frac{h+\rho k}{k\sqrt{1-\rho^2}}\right )-\eta \right].
\end{align}
Summarizing the two cases, we get the closed form expression of $F_{\widetilde{q}}$:
\begin{align}
    F_{\widetilde{q}}(x)
    &=Z^{-1}\Bigg[\frac{1}{2}\Phi(h)-yT\left(h,\frac{k+\rho h}{h\sqrt{1-\rho^2}}\right )+\frac{y}{2}\Phi(k) -yT\left(k,\frac{h+\rho k}{k\sqrt{1-\rho^2}}\right )+y\eta \Bigg]
    \\
    &=Z^{-1}\Bigg[\frac{1}{2}\Phi(h)-yT\left(h,\frac{k}{h\sqrt{1-\rho^2}}+\sigma\right )+\frac{y}{2}\Phi(k) -yT\left(k,\frac{h}{k\sqrt{1-\rho^2}}+\sigma\right )+y\eta \Bigg].
\end{align}
Provided the above, the optimal $\sigma^\star$ can be computed by numerical integration of Eqn~\eqref{eqn:sigmaintegral}. For special cases, we provide additional formulas:
\begin{align}
    &(1)\, x = \mu,~\mu \neq 0: F_{\widetilde{q}}(x)= Z^{-1} \left[\frac{1}{4}-\frac{y\mathrm{sign}(k)}{4}+\frac{y}{2}\Phi(k)-yT(k,\sigma)+y\eta \right ];
    \\
    &(2)\, x \neq \mu,~\mu = 0: F_{\widetilde{q}}(x)= 2 \left [\frac{1}{2}\Phi(h)-yT(h,\sigma)+\frac{y}{4}-\frac{ y \mathrm{sign}(h)}{4}+y\eta \right ];
    \\
    &(3)\, x = \mu,~\mu = 0: F_{\widetilde{q}}(x)= \frac{1}{2}-\frac{y}{ \pi} \mathrm{arctan}(\sigma).
\end{align}

\subsection{Square Link Function for Poisson Regression}
\label{appx:square_poisson_likelihood}

Consider Poisson regression, which uses the Poisson likelihood $p(y|g) = g^y \exp(-g)/y!$ to model count data $y\in \mathbb N$, with the square link function $g(f) = f^2$ \citep{walder2017fast,flaxman2017poisson}. We use the square link function because it is more mathematically convenient than the exponential function. Given the cavity distribution $q^{\setminus i}(f) = \mN(f|\mu,\sigma^2)$, we want the tilted distribution $\widetilde{q}(f) = q^{\setminus i}(f)p(y|g(f))/Z$ where the normalizer $Z$ is derived as:
\begin{align}
    Z &= \int_{-\infty}^{\infty} q^{\setminus i}(f)p(y|g) \intd f
    \\
    &= \int_{-\infty}^{\infty} \frac{1}{\sqrt{2\pi \sigma^2}} \exp\left (-\frac{(f-\mu)^2}{2\sigma^2} \right ) f^{2y}\exp(-f^2)/y! \intd f
    \\
    &\overset{\text{(a)}}{=} \frac{1}{\sqrt{2\pi \sigma^2}y!\exp(\mu^2/(1+2\sigma^2))}\int_{-\infty}^{\infty} f^{2y}  \exp\left (-\frac{(f-\mu/(1+2\sigma^2))^2}{2\sigma^2/(1+2\sigma^2)} \right ) \intd f
    \\
    &\overset{\text{(b)}}{=} \frac{\left(\frac{2\sigma^2}{1+2\sigma^2}\right)^{y+\frac{1}{2}}}{\sqrt{2\pi \sigma^2}y!\exp(\mu^2/(1+2\sigma^2))}\Gamma
    \left(y+\frac{1}{2}\right) \,
    _1F_1\left(-y;\frac{1}{2};-\frac{\mu ^2}{2 \sigma
    ^2 (1+2\sigma^2)}\right)
    \\
    &= \frac{\alpha^{y+\frac{1}{2}}}{\sqrt{2\pi \sigma^2}y!\exp(h)}\Gamma
    \left(y+\frac{1}{2}\right) \,
    _1F_1\left(-y;\frac{1}{2};-\frac{h}{2 \sigma
    ^2}\right),
    \\
    \alpha &= \frac{2\sigma^2}{1+2\sigma^2}, ~~ h = \frac{\mu^2}{1+2\sigma^2} \label{eq:Z_square_poisson}
\end{align}
where the step (a) rewrites the product of two exponential functions into the form of the Gaussian distribution, (b) is achieved through Mathematica \citep{Mathematica}, $\Gamma(\cdot)$ is the Gamma function and $_1F_1\left(-y;\frac{1}{2};-\frac{h^2}{2 \sigma
^2 }\right)$ is the confluent hypergeometric function of the first kind. Furthermore, we compute the first derivative of $\log Z$ w.r.t. $\mu$ and then the mean of the tilted distribution:
\begin{align}
    \partial_{\mu} \log Z &=\left(\frac{y\,
    _1F_1\left(-y+1;\frac{3}{2};-\frac{h}{2 \sigma
    ^2}\right)}{\sigma^2 \,
    _1F_1\left(-y;\frac{1}{2};-\frac{h}{2 \sigma
    ^2}\right)}-1 \right)\frac{2\mu}{1+2\sigma^2}
    \\
    \Longrightarrow \mu_{\widetilde{q}} &= \sigma^2 \partial_{\mu} \log Z +\mu.
    \\
    \partial^2_{\mu} \log Z &=\left(\frac{y\,
    _1F_1\left(-y+1;\frac{3}{2};-\frac{h}{2 \sigma
    ^2}\right)}{\sigma^2\,
    _1F_1\left(-y;\frac{1}{2};-\frac{h}{2 \sigma
    ^2}\right)}-1 \right)\frac{2}{1+2\sigma^2}-
    \\
    &\quad \left(\frac{2(1-y)\,
    _1F_1\left(-y+2;\frac{5}{2};-\frac{h}{2 \sigma
    ^2}\right)}{3\,
    _1F_1\left(-y;\frac{1}{2};-\frac{h}{2 \sigma
    ^2}\right)}+
    \frac{2y \,
    _1F_1\left(-y+1;\frac{3}{2};-\frac{h}{2 \sigma
    ^2}\right)^2}{\,
    _1F_1\left(-y;\frac{1}{2};-\frac{h}{2 \sigma
    ^2}\right)^2} \right)\frac{2\mu^2 y}{\sigma^4(1+2\sigma^2)^2}
    \\
    \Longrightarrow \sigma^2_{\widetilde{q}} &= \sigma^4 \partial_{\mu}^2 \log Z + \sigma^2
\end{align}
Finally, we derive the CDF of the tilted distribution $\widetilde{q}$ by using the binomial theorem:
\begin{align}
    F_{\widetilde{q}}(x) &= Z^{-1}\int_{-\infty}^x p(y|g) \mN(f|\mu,\sigma^2) \intd f
    \\
    &\overset{\text{(a)}}{=} A \int_{-\infty}^x  f^{2y}  \exp\left (-\frac{(f-\mu/(1+2\sigma^2))^2}{2\sigma^2/(1+2\sigma^2)} \right ) \intd f
    \\
    &= A\int_{-\infty}^{x-\frac{\mu}{1+2\sigma^2}}  \left(f+\frac{\mu}{1+2\sigma^2}\right)^{2y}  \exp\left (-\frac{f^2}{2\sigma^2/(1+2\sigma^2)} \right ) \intd f
    \\
    &\overset{\text{(b)}}{=} A\int_{-\infty}^{x-\beta} \left[\sum_{k=0}^{2y} \begin{pmatrix}
        2y \\ k
    \end{pmatrix} f^k \beta^{2y-k} \right]\exp\left (-\frac{f^2}{\alpha} \right ) \intd f 
    \\
    &= A \sum_{k=0}^{2y} \begin{pmatrix}
        2y \\ k
    \end{pmatrix} \beta^{2y-k}\left[\int_{-\infty}^{0}f^k  \exp\left (-\frac{f^2}{\alpha} \right ) \intd f+\int_{0}^{x-\beta}f^k  \exp\left (-\frac{f^2}{\alpha} \right ) \intd f\right]
    \\
    &\overset{\text{(c)}}{=} \frac{A}{2}\sum_{k=0}^{2y} \begin{pmatrix}
        2y \\ k
    \end{pmatrix} \beta^{2y-k}\alpha ^{\frac{k+1}{2}}\left[ (-1)^k \Gamma
    \left(\frac{k+1}{2}\right)+ 
    \text{sgn}(x-\beta)^{k+1} \left(\Gamma
    \left(\frac{k+1}{2}\right)-\Gamma
    \left(\frac{k+1}{2},\frac{(x-\beta)^2}{\alpha }\right)\right)\right]
    \\
    A&=\frac{Z^{-1}}{\sqrt{2\pi \sigma^2}y!\exp(\mu^2/(1+2\sigma^2))}=\left[\alpha^{y+\frac{1}{2}}\Gamma
    \left(y+\frac{1}{2}\right) \,
    _1F_1\left(-y;\frac{1}{2};-\frac{h}{2 \sigma
    ^2}\right)\right]^{-1}, ~~ \beta = \frac{\mu}{1+2\sigma^2},
\end{align}
where the step (a) has been derived in (a) of Eqn.~\eqref{eq:Z_square_poisson}, (b) applies the binomial theorem and (c) is obtained through Mathematica \citep{Mathematica}. And, the function $\Gamma(a,z)=\int_z^{\infty}t^{a-1}e^{-t} \intd t$ is the upper incomplete gamma function and $\text{sgn}(x)$ is the sign function, equaling $1$ when $x>0$, $0$ when $x=0$ and $-1$ when $x<0$.

\section{Proof of Convexity}
\label{sec:proof_convexity}
\paragraph{Theorem} Given two probability measures in $\M_+^1(\mathbb{R})$: a Gaussian $\mN(\mu, \sigma^2)$ with  mean $\mu$ and  standard deviation $\sigma>0$, and an arbitrary measure $\widetilde{q}$, the $L_p$ WD $\W_p^p(\widetilde{q},\mN)$ is strictly convex about $\mu$ and $\sigma$.

\begin{prf}
Let $F_{\widetilde{q}}^{-1}(y)$ and $F_{\mN}^{-1}(y)=\mu + \sqrt{2}\sigma\erf^{-1}(2y-1)$, $y \in [0,1]$, be \revA{the} quantile functions of $\widetilde{q}$ and the Gaussian $\mN$, where erf is the error function. 
Then, we consider two distinct Gaussian measures $\mN(\mu_1, \sigma_1^2)$ and $\mN(\mu_2, \sigma_2^2)$ and a convex combination w.r.t.~their parameters $\mN(a_1 \mu_1 + a_2 \mu_2, (a_1 \sigma_1 + a_2 \sigma_2)^2)$ with $a_1,a_2\in \mathbb R_{+}$ and $a_1 + a_2 = 1$. Given the above, we further define $\varepsilon_k(y) = F_{\widetilde{q}}^{-1}(y)-\mu_k-\sigma_k\sqrt{2}\erf^{-1}(2y-1)$, $k=1,2$, for notational simplification, and derive the convexity as:
\begin{align}
     &\W_p^p(\widetilde{q},\mN(a_1 \mu_1 + a_2 \mu_2,(a_1 \sigma_1 + a_2 \sigma_2)^2))
    \overset{(a)}{=}\int_{0}^{1} |a_1\varepsilon_1(y)+a_2 \varepsilon_2(y)|^p \intd y\overset{(b)}{\leq} \int_{0}^{1} (a_1|\varepsilon_1(y)|+
    \\
    &\qquad \qquad a_2 |\varepsilon_2(y)| )^p \intd y \overset{(c)}{\leq} 
      a_1 \W_p^p(\widetilde{q},\mN(\mu_1,\sigma_1^2))+a_2 \W_p^p(\widetilde{q},\mN(\mu_2,\sigma_2^2)),
\end{align}
where steps (a), (b) and (c) are obtained by applying \sref{Proposition}{prop:1d}, non-negativity of the absolute value, and the convexity of $f(x)=x^p$, $p\geq 1$, over $\mathbb R_+$ respectively. 
The equality at $(b)$ holds iff 
$\varepsilon_k(y)\geq 0, k=1,2, \forall y \in [0,1]$, and $(c)$'s equality
holds iff $|\varepsilon_1(y)| = |\varepsilon_2(y)|$, $\forall y \in [0,1]$.
These two conditions for equality can't be attained simultaneously as otherwise \revA{it would contradict} that $\mN(\mu_1,\sigma^2_1)$ is different from $\mN(\mu_2,\sigma^2_2)$. So, $\W_{p}^p(\widetilde{q},\mN)$, $p \geq 1$, is strictly convex about $\mu$ and $\sigma$.
\end{prf}

\section{Proof of Variance Difference}
\label{sec:proof_variance_diff}
\paragraph{Theorem} The variance of the Gaussian approximation to a univariate tilted distribution $\widetilde{q}(f)$ as estimated by QP and EP satisfy $\sigma_{\text{QP}}^2\leq  \sigma^2_{\text{EP}}$. \label{thm:variance}
\begin{prf}
Let $\mN(\mu_{\text{QP}},\sigma^2_{\text{QP}})$ be the optimal Gaussian in QP. As per \sref{Proposition}{prop:1d}, we reformulate the $L_2$ WD based projection $\W_2^2(\widetilde{q},\mN(\mu_{\text{QP}},\sigma_{\text{QP}}^2))$ w.r.t. quantile functions:
{\medmuskip=1mu
	\thinmuskip=1mu
	\thickmuskip=1mu
\begin{align}
    \W_2^2(\widetilde{q},\mN(\mu_{\text{QP}},&\sigma_{\text{QP}}^2)) 
    = \int_{0}^{1}|F_{\widetilde{q}}^{-1}(y)-\mu_{\text{QP}}-\sqrt{2}\sigma_{\text{QP}} \erf^{-1}(2y-1)|^2 \intd y
    = \int_{0}^{1}\underbrace{(F_{\widetilde{q}}^{-1}(y)-\mu_{\text{QP}})^2}_{\sigma_{\text{EP}}^2}
    \\
    +&\underbrace{(\sqrt{2}\sigma_{\text{QP}}\erf^{-1}(2y-1))^2}_{\sigma_{\text{QP}}^2}-\underbrace{2(F_{\widetilde{q}}^{-1}(y)-\mu_{\text{QP}})\sqrt{2}\sigma_{\text{QP}} \erf^{-1}(2y-1)}_{\mathrm{(A)}} \intd y=\sigma_{\text{EP}}^2-\sigma^{2}_{\text{QP}},
\end{align}}where for (A), we used $\int \mu_{\text{QP}}\sigma_{\text{QP}} \erf^{-1}(2y-1)\intd y = 0$ and the remaining factor can be easily shown to be equal to $2\sigma_{\text{QP}}^2$. Furthermore, due to the non-negativity of the WD, we have $\sigma^{2}_{\text{EP}} \geq \sigma^{2}_{\text{QP}}$, and the equality holds iff $\widetilde{q}$ is Gaussian.
\end{prf}

\section{Proof of Locality Property}
\label{sec:proof_locality}
\paragraph{Theorem}
Minimization of $\W_2^2 (\widetilde{q}(\bm f), \mN (\bm f))$ w.r.t. $\mN(\bm f)$ results in $q^{\setminus i}(\bm f_{\setminus i}|f_i)=\mN(\bm f_{\setminus i}|f_i)$.
\begin{prf}
We first apply the decomposition of the $L_2$ norm to rewriting the $\W_2^2(\widetilde{q}(\bm f), \mN(\bm f))$ as below (see detailed derivations in  \sref{Appendix}{appx:details_of_eq_wd_tildeq_N}):
{\medmuskip=1mu
	\thinmuskip=1mu
	\thickmuskip=1mu
\begin{align}
&\W_2^2\left (\widetilde{q},\mathcal{N}\right )=
\inf_{\pi_i 
} \EPT_{\pi_i}\Big[ \|  f_{i}-f'_{i}\|_2^2 +\W_2^2(q^{\setminus i}_{\setminus i|i},\mN_{\setminus i|i})\Big],~~~~~\label{eq:wd_tildeq_N}
\end{align}}where the prime indicates that the variable is from the Gaussian $\mN$, and for simplification, we use the notation $\pi_i$ for the joint distribution $\pi(f_i, f'_i)$ which belongs to a set of measures $U(\widetilde{q}_i,\mN_i)$. 
Since $q^{\setminus i}(\bm f)$ is known to be Gaussian, we define it in a partitioned form:
\begin{align}\label{eq:cavity_definition}
    q^{\setminus i}(\bm f) \equiv \mN \left(\begin{bmatrix} \bm f_{\setminus i} \\
    f_i
    \end{bmatrix}
    \;\middle|\;
    \begin{bmatrix} \bm m_{\setminus i} \\
    m_i
    \end{bmatrix}, 
    \begin{bmatrix} \bm S_{\setminus i} & \bm S_{\setminus ii}\\
    \bm S_{\setminus ii}^\T & S_{i}
    \end{bmatrix}\right),
\end{align}
and the conditional $q^{\setminus i}(\bm f_{\setminus i}| f_i)$ is expressed as:
\begin{align}
    q^{\setminus i}(\bm f_{\setminus i}|f_i) = \mN(\bm f_{\setminus i}|\bm m_{\setminus i|i}, \bm S_{\setminus i|i}), ~~&\bm m_{\setminus i|i}= \bm m_{\setminus i}+\bm S_{\setminus ii} S_i^{-1}(f_i-m_i) \equiv \bm a f_i+\bm b, \label{eq:condtional_mean_q}
    \\
    &\bm S_{\setminus i|i} = \bm S_{\setminus i}-\bm S_{\setminus ii}S_i^{-1}\bm S_{\setminus ii}^\T.\label{eq:condtional_covariance_q}
\end{align}
We define a similar partioned expression for the Gaussian $\mN(\bm f')$ by adding primes to variables and parameters on the r.h.s. of \eqnref{eq:cavity_definition}, and as a result, the conditional $\mN(\bm f_{\setminus i}'| f_i')$ is written as:
\begin{align}
    \mN(\bm f_{\setminus i}'|f_i')=\mN(\bm m_{\setminus i|i}', \bm S_{\setminus i|i}'),~~
    &\bm m_{\setminus i|i}' = \bm m_{\setminus i}'+\bm S_{\setminus ii}'S_i^{\prime-1}(f_i'-m_i') \equiv \bm a'f_i'+\bm b', \label{eq:condtional_mean_N}
    \\
    &\bm S_{\setminus i|i}'=\bm S'_{\setminus i}-\bm S'_{\setminus ii} S_{i}^{\prime-1}\bm S_{\setminus ii}^{\prime~\T}.\label{eq:condtional_covariance_N}
\end{align}
Given the above definitions, we exploit \sref{Proposition}{prop:translation} to take the means out of the $L_2$ WD on the r.h.s. of \eqnref{eq:wd_tildeq_N}:
\begin{align}
\W_2^2\left (\widetilde{q},\mathcal{N}\right )
&=\inf_{\pi_i
} \EPT_{\pi_i}\Big[ \| f_{i}-f_i'\|_2^2 + \|\bm m_{\setminus i|i}-\bm m_{\setminus i|i}'\|_2^2 \Big]
 +\underbrace{\W_2^2\left(\mN(\bm 0, \bm S_{\setminus i|i}),\mN(\bm 0, \bm S_{\setminus i|i}')\right)}_{\mathrm{(A)}}.~~
\label{eq:minimize0}
\end{align}
Minimizing this function requires optimizing $m_i'$, $\bm m_{\setminus i}'$, $S_i'$, $\bm S_{\setminus i}'$ and $\bm S_{\setminus ii}'$. As $\bm S_{\setminus i}'$ is only contained in $\bm S_{\setminus i | i}$ and isolated into the term $\mathrm{(A)}$,  it can be optimized by simply setting
{\medmuskip=1mu
	\thinmuskip=1mu
	\thickmuskip=1mu
\begin{align}
\bm S_{\setminus i|i}'&=\bm S_{\setminus i|i} \label{eq:optimal_S_setminusi_i}
\overset{\text{Eqn.~\eqref{eq:condtional_covariance_N}}}{\Longrightarrow}\bm S_{\setminus i}^{(n)*}=\bm S_{\setminus i|i}+\bm S'_{\setminus ii}S_{i}^{\prime -1} \bm S_{\setminus ii}^{\prime~\T}.
\end{align}}As a result, $\mathrm{(A)}$ is minimized to zero. Next, we plug in expressions of $\bm m_{\setminus i|i}$ and $\bm m_{\setminus i|i}'$ (\eqnref{eq:condtional_mean_q} and \eqnref{eq:condtional_mean_N}) into optimized \eqnref{eq:minimize0}:
{\medmuskip=0mu
	\thinmuskip=0mu
	\thickmuskip=0mu
\begin{align}
\min_{\bm S_{\setminus i}'} \text{\eqref{eq:minimize0}}
&=\inf_{\pi_i
} \EPT_{\pi_i}\left[ \| f_i-f_i'\|_2^2 + \|\bm a f_i-\bm a' f_i'+\bm b-\bm b'\|_2^2\right],~~~~
\label{eq:minimize1}
\end{align}} where $\bm m_{\setminus i}'$ is only contained by $\bm b'$. Thus, we can optimize it by zeroing the derivative of the above function  about $\bm m_{\setminus i}'$, which results in:
\begin{align}
&\bm b'=\bm b + \bm a \mu_{\widetilde{q}_i}-\bm a' m_i'\overset{\text{Eqn.~\eqref{eq:condtional_mean_N}}}{\Longrightarrow}\label{eq:optimal_b}
\bm m_{\setminus i}^{(n)*} =\bm S_{\setminus ii}' S_i^{\prime-1} m_i'+ \bm b + \bm a \mu_{\widetilde{q}_i}-\bm a' m_i',
\end{align}
where $ \mu_{\widetilde{q}_i}$ is the mean of $\widetilde{q}(f_i)$. The minimum value of \eqnref{eq:minimize1} thereby is (see details in \autoref{appx:details_of_minimize2}):
\begin{align}
&\min_{\bm m_{\setminus i}'} \text{\eqref{eq:minimize1}}
 = (1+\bm a^\T \bm a') \W_2^2(\widetilde{q}_i,\mN_i)+\|\bm a\|_2^2 \sigma_{\widetilde{q}_i}^2+
 \|\bm a'\|_2^2 S_i'- \bm a^{\T}\bm a'\Big[\sigma_{\widetilde{q}_i}^2+S_i'+( \mu_{\widetilde{q}_i}-m_i')^2\Big]~~~
\label{eq:minimize2}
\end{align}
where $\sigma_{\widetilde{q}_i}^2$ is the variance of $\widetilde{q}(f_i)$. This function can be further simplified using the quantile based reformulation of $\W_2^2(\widetilde{q}_i,\mN_i)$ (see details in \sref{Appendix}{appx:details_wd_in_proof}) which results in:
{\medmuskip=1.1mu
	\thinmuskip=1.1mu
	\thickmuskip=1.1mu
\begin{align}
\text{\eqref{eq:minimize2}}&= \W_2^2(\widetilde{q}_i,\mN_i)+\|\bm a\|_2^2 \sigma^2_{\widetilde{q}_i}
\underbrace{-
 2^{\frac{3}{2}}\bm a^\T \bm a'c_{\widetilde{q}_i}S_i'^{\frac{1}{2}}+\|\bm a'\|_2^2 S_i'}_{\mathrm{(B)}}.~~~~
\label{eq:minimize3}
\end{align}}
Now, we are left with optimizing  $m_i'$,  $S_i'$ and $\bm S_{\setminus ii}'$. To optimize $\bm S_{\setminus ii}'$,  which only exists in the above term $\mathrm{(B)}$, we zero the derivative of $\mathrm{(B)}$ w.r.t. $\bm S_{\setminus ii}'$ and this yields:
\begin{align}
\bm a^{\prime *} &= 2^{\frac{1}{2}}(S_i')^{-\frac{1}{2}}c_{\widetilde{q}_i}\bm a \label{eq:optimal_a}
\overset{\text{Eqn.~\eqref{eq:condtional_mean_N}}}{\Longrightarrow}\bm S_{\setminus ii}^{\prime *} = (2S_i')^{\frac{1}{2}}c_{\widetilde{q}_i}\bm a,~~~~
\end{align} and the minimum value of \eqnref{eq:minimize3} is
\begin{align}\label{eq:final_obj}
\min_{\bm S_{\backslash ii}'} \text{\eqref{eq:minimize3}}= \W_2^2(\widetilde{q}_i,\mN_i)+\|\bm a\|_2^2 (\sigma^2_{\widetilde{q}_i}-2c_{\widetilde{q}_i}^2).~~~~
\end{align}
The results of optimizing $m_i'$ and $S_i'$ in the above equation have already been provided in \eqnref{eq:optimal_sigma}: $m_i^{\prime *} = \mu_{\widetilde{q}_i}$ and $S_i^{\prime *}=2c_{\widetilde{q}_i}^2$.
By plugging them into \eqnref{eq:optimal_a} and \eqnref{eq:optimal_b},  we have $\bm a^{\prime *}=\bm a$ and $\bm b^{\prime *}=\bm b$. Finally, using \eqnref{eq:optimal_S_setminusi_i}, we obtain $q^{\setminus i}(\bm f_{\setminus i}|f_i)=\mN(\bm f_{\setminus i} | \bm a f_i + \bm b, \bm S_{\backslash i | i})=\mN(\bm f_{\setminus i} | \bm a' f_i + \bm b', \bm S_{\backslash i | i}')=\mN(\bm f_{\setminus i}|f_i)$ , which concludes the proof.
\end{prf}
\subsection{Details of Eqn.~\texorpdfstring{\eqref{eq:KL_locality}}{}} \label{appx:details_decomposition_KL_proof_EP}
\begin{align}
    \KL(\widetilde{q}(\bm f) \| \mN(\bm f)) &= \int \widetilde{q}(\bm f) \log \frac{\widetilde{q}(\bm f_{\setminus i}|f_i)\widetilde{q}(f_i) }{\mN(\bm f_{\setminus i}|f_i)\mN(f_i)}  \intd \bm f
    \\
    &=\int \widetilde{q}(f_i) \log \frac{\widetilde{q}(f_i) }{\mN(f_i)} \intd f_i+\int \widetilde{q}(f_i) \int \widetilde{q}(\bm f_{\setminus i}|f_i)\log \frac{\widetilde{q}(\bm f_{\setminus i}|f_i)}{\mN(\bm f_{\setminus i}|f_i)} \intd \bm f_{\setminus i} \intd f_i
    \\
    &=\KL\big(\widetilde{q}(f_i)\| \mN(f_i)\big )+\EPT_{\widetilde{q}(f_i)}\Big[\KL\big(\widetilde{q}(\bm f_{\setminus i}|f_i) \| \mN(\bm f_{\setminus i}|f_i)\big)\Big]
    \\
    \widetilde{q}(\bm f_{\setminus i}|f_i)&=\frac{\widetilde{q}(\bm f)}{\widetilde{q}(f_i)} \propto \frac{p(\bm f )\cancel{p(y_i|f_i)}\prod_{j\neq i} t_j(\bm f) }{q^{\setminus i}(f_i)\cancel{p(y_i|f_i)}}
     \\
     &\qquad \quad ~~ = q^{\setminus i}(\bm f_{\setminus i}|f_i) \label{eq:tilted_eq_cavity}.
\end{align}

\subsection{Details of Eqn.~\texorpdfstring{\eqref{eq:wd_tildeq_N}}{}}
\label{appx:details_of_eq_wd_tildeq_N}
\begin{align}
\W_2^2\left (\widetilde{q}(\bm f),\mathcal{N}(\bm f)\right ) 
&\equiv \inf_{\pi \in U(\widetilde{q},\mN)} \EPT_{\pi} \left( \| \bm f-\bm f'\|_2^2 \right ) 
\\
&=\inf_{\pi \in U(\widetilde{q},\mN)} \EPT_{\pi}\left(\| f_{i}-f'_i\|_2^2 \right )  + \EPT_{\pi} \left( \| \bm f_{\setminus i}-\bm f'_{\setminus i}\|_2^2 \right )
\\
&\overset{\text{(a)}}{=}\inf_{\pi \in U(\widetilde{q},\mN)} \EPT_{\pi_i}\Big[ \| f_{i}- f'_i\|_2^2 +\EPT_{\pi_{\setminus i|i}} \left( \| \bm f_{\setminus i}-\bm f'_{\setminus i}\|_2^2 \right )\Big]
\\
&\overset{\text{(b)}}{=}
\inf_{\pi_i 
} \EPT_{\pi_i}\Big[ \| f_{i}- f'_{i}\|_2^2 +\inf_{\pi_{\setminus i | i}
} \EPT_{\pi_{\setminus i|i}} \left( \| \bm f_{\setminus i}-\bm f'_{\setminus i}\|_2^2 \right )\Big]
\\
&=
\inf_{\pi_i 
} \EPT_{\pi_i}\Big[ \|  f_{i}-f'_{i}\|_2^2 +\W_2^2(\widetilde{q}_{\setminus i|i},\mN_{\setminus i|i})\Big]
\\
&\overset{(\text{c})}=
\inf_{\pi_i 
} \EPT_{\pi_i}\Big[ \|  f_{i}-f'_{i}\|_2^2 +\W_2^2(q^{\setminus i}_{\setminus i|i},\mN_{\setminus i|i})\Big],
\end{align}where the superscript prime indicates that the variable is from the Gaussian $\mN$.
In (a), $\pi_i=\pi(f_i, f'_i)$ and $\pi_{\setminus i | i}=\pi(\bm f_{\setminus i}, \bm f'_{\setminus i}|f_i, f'_i)$. In (b), the first and the second $\inf$ are over $U(\widetilde{q}_i,\mN_i)$ and  $U(\widetilde{q}_{\setminus i|i},\mN_{\setminus i|i})$ respectively. (c) is due to $\widetilde{q}(\bm f_{\setminus i}|f_i)$ being equal to $q^{\setminus i}(\bm f_{\setminus i} | f_i)$ (refer to Eqn.~\eqref{eq:tilted_eq_cavity}).

\subsection{Details of Eqn.~\texorpdfstring{\eqref{eq:minimize2}}{}}
\label{appx:details_of_minimize2}
\begin{align}
&\min_{\bm m_{\setminus i}'} \text{Eqn.~\eqref{eq:minimize1}}
\\
&=\inf_{\pi_i
} \EPT_{\pi_i}\Big[ \| f_i-f_i'\|_2^2 + \|\bm a (f_i-\mu_{\widetilde{q}_i})- \bm a' (f_i'-m_i')\|_2^2\Big]
\\
& =\inf_{\pi_i
} \EPT_{\pi_i}\Big[ \| f_i-f_i'\|_2^2\Big] +\|\bm a\|_2^2 \sigma_{\widetilde{q}_i}^2 +\|\bm a'\|_2^2 S_i'-2\bm a^{\T}\bm a'\EPT_{\pi_i}\Big(f_i f_i'- \mu_{\widetilde{q}_i} m_i' \Big)
\\
& =\inf_{\pi_i
} \EPT_{\pi_i}\Big[ \| f_i-f_i'\|_2^2\Big] +\|\bm a\|_2^2 \sigma_{\widetilde{q}_i}^2 +\|\bm a'\|_2^2 S_i'+\bm a^{\T}\bm a'\EPT_{\pi_i}\Big(\|f_i-f_i'\|_2^2 - f_i^2-(f_i')^2 +2\mu_{\widetilde{q}_i} m_i'\Big)
\\
& =\inf_{\pi_i
} \EPT_{\pi_i}\Big[ \| f_i-f_i'\|_2^2\Big] +\|\bm a\|_2^2 \sigma_{\widetilde{q}_i}^2 +\|\bm a'\|_2^2 S_i' + \bm a^{\T}\bm a'\EPT_{\pi_i}\Big(\|f_i-f_i'\|_2^2 - (f_i-\mu_{\widetilde{q}_i})^2-
\\
&\quad 2f_i\mu_{\widetilde{q}_i}+\mu_{\widetilde{q}_i}^2-(f_i'-m_i')^2-2f_i'm_i'+(m_i')^2 +2\mu_{\widetilde{q}_i} m_i'\Big)
\\
& =(1+\bm a^\T \bm a') \W_2^2(\widetilde{q}_i,\mN_i) +\|\bm a\|_2^2 \sigma_{\widetilde{q}_i}^2 +\|\bm a'\|_2^2 S_i'-\bm a^{\T}\bm a'\Big(\sigma_{\widetilde{q}_i}^2+\mu_{\widetilde{q}_i}^2+S_i'+(m_i')^2-2\mu_{\widetilde{q}_i} m_i'\Big)
\\
& = (1+\bm a^\T \bm a') \W_2^2(\widetilde{q}_i,\mN_i)+\|\bm a\|_2^2 \sigma_{\widetilde{q}_i}^2+\|\bm a'\|_2^2 S_i'-\bm a^{\T}\bm a'\Big[\sigma_{\widetilde{q}_i}^2+S_i'+( \mu_{\widetilde{q}_i}-m_i')^2\Big]
\end{align}

\subsection{Details of Eqn.~\texorpdfstring{\eqref{eq:minimize2}}{}}
\label{appx:details_wd_in_proof}
We first use Proposition \ref{prop:1d} to reformulate the $L_2$ WD $\W_{2}^2(\widetilde{q}_i, \mN_i)$ as:
\begin{align}
    \W_{2}^2(\widetilde{q}_i, \mN_i) &= \int_{0}^{1} \big(F_{\widetilde{q}_i}^{-1}(y)-m_i'-\sqrt{2S_i'}\erf^{-1}(2y-1)\big )^2 \intd y,
    \\
    &=\int_{0}^{1} (F_{\widetilde{q}_i}^{-1}(y)-m_i')^2+2S_i'\erf^{-1}(2y-1)^2-2\sqrt{2S_i'}\erf^{-1}(2y-1)(F_{\widetilde{q}_i}^{-1}(y)-m_i') \intd y,
    \\
    &=\int_{0}^{1} (F_{\widetilde{q}_i}^{-1}(y)-\mu_{\widetilde{q}_i}+\mu_{\widetilde{q}_i}-m_i')^2\intd y+S_i'-2\sqrt{2S_i'}c_{\widetilde{q}_i} ,
    \\
    &= \sigma^2_{\widetilde{q}_i} +(\mu_{\widetilde{q}_i} -m_i')^2+S_i'-2c_{\widetilde{q}_i}\sqrt{2S_i'},
\end{align}
where $F_{\widetilde{q}_i}^{-1}(y)$ is the quantile function of $\widetilde{q}(f_i)$ and $c_{\widetilde{q}_i}\equiv \int_0^{1}F_{\widetilde{q}_i}^{-1}(y)\erf^{-1}(2y-1) \intd y$. Next, we plug this reformulation into Eqn.~\eqref{eq:minimize2}:
\begin{align}
    \text{Eqn.~\eqref{eq:minimize2}}&=
    \W_2^2(\widetilde{q}_i,\mN_i)+\bm a^\T \bm a' \W_2^2(\widetilde{q}_i,\mN_i)+\|\bm a\|_2^2 \sigma_{\widetilde{q}_i}^2+\|\bm a'\|_2^2 S_i'- \bm a^{\T}\bm a'\Big[\sigma_{\widetilde{q}_i}^2+S_i'+( \mu_{\widetilde{q}_i}-m_i')^2\Big]
    \\
    &=\W_2^2(\widetilde{q}_i,\mN_i)+\bm a^\T \bm a'\left[\cancel{\sigma^2_{\widetilde{q}_i} +(\mu_{\widetilde{q}_i} -m_i')^2+S_i'}-2c_{\widetilde{q}_i}\sqrt{2S_i'}\right]+\|\bm a\|_2^2 \sigma_{\widetilde{q}_i}^2+\|\bm a'\|_2^2 S_i'
    \\
    &\quad - \bm a^{\T}\bm a'\Big[\cancel{\sigma_{\widetilde{q}_i}^2+S_i'+( \mu_{\widetilde{q}_i}-m_i')^2}\Big]
    \\
    &=\W_2^2(\widetilde{q}_i,\mN_i)-2c_{\widetilde{q}_i}\sqrt{2S_i'}\bm a^\T \bm a'+\|\bm a\|_2^2 \sigma_{\widetilde{q}_i}^2+\|\bm a'\|_2^2 S_i'
\end{align}

\section{More Details of EP}
\label{appx:EP_details}
We use the expressions $\widetilde{q}(\bm f) = q^{\setminus i}(\bm f) p(y_i|f_i)/Z_{\widetilde{q}}$ and $q^{\setminus i}(\bm f) = q(\bm f)/(t_i(f_i)Z_{q^{\setminus i}})$, and the derivation of $\KL(\widetilde{q}(\bm f) \| q(\bm f))=\KL(\widetilde{q}(f_i) \| q(f_i))$ is shown as below:
\begin{align}
    \KL(\widetilde{q}(\bm f) \| q(\bm f)) &= \int \widetilde{q}(\bm f) \log \frac{q^{\setminus i}(\bm f) p(y_i|f_i)}{Z_{\widetilde{q}} q(\bm f)} \intd \bm f
    \\
    &= \int \widetilde{q}(\bm f) \log \frac{\cancel{q(\bm f)} p(y_i|f_i)}{Z_{q^{\setminus i}}Z_{\widetilde{q}}\cancel{q(\bm f)}t_i(f_i)} \intd \bm f
    \\
    & = \int \widetilde{q}(f_i) \log \frac{p(y_i|f_i)}{Z_{q^{\setminus i}}Z_{\widetilde{q}}t_i(f_i)} \intd f_i
    \\
    & = \int \widetilde{q}(f_i) \log \frac{q^{\setminus i}(f_i)p(y_i|f_i)}{Z_{q^{\setminus i}}Z_{\widetilde{q}}q^{\setminus i}(f_i)t_i(f_i)} \intd f_i
    \\
    & = \int \widetilde{q}(f_i) \log \frac{\widetilde{q}(f_i)}{q(f_i)} \intd f_i
    \\
    &= \KL(\widetilde{q}(f_i) \| q(f_i))
\end{align}

\section{Predictive Distributions of Poisson Regression}
\label{appx:square_poisson_pred}

Given the approximate predictive distribution $f(\bm x_*) = \mN(\mu_*, \sigma_*^2)$ and the relation $g(f) = f^2$, it is straightforward to derive the corresponding $g(\bm x_*)\sim \text{Gamma}(k_*,c_*)$\footnote{$\text{Gamma}(x|k,c) = \frac{1}{\Gamma(k)c^k}x^{k-1}e^{-x/c}$.} where the shape $k_*$ and the scale $c_*$ are expressed as \citep{walder2017fast,zhang2020variational}:
\begin{align}
    k_*=\frac{(\mu_*^2+\sigma_*^2)^2}{2\sigma^2_*(2\mu_*^2+\sigma_*^2)}, ~~ c_* = \frac{2\sigma^2_*(2\mu^2_*+\sigma^2_*)}{\mu^2_*+\sigma^2_*}.
\end{align}
Furthermore, the predictive distribution of the count value $y \in \mathbb N$ can also be derived straightforwardly:
\begin{align}
    p(y) &= \int_{0}^{\infty} p(g_*) p(y|g_*) \intd g_*
    \\
    &= \int \text{Gamma}(g_* | k_*,c_*) \text{Poisson}(y|g_*) \intd g_*
    \\
    &=\frac{c_*^y (c_*+1)^{-k_*-y} \Gamma
    (k_*+y)}{y! \Gamma (k_*)} = \text{NB}(y|k_*,c_*/(1+c_*)),
\end{align}
where $g_* = g(\bm x_*)$ and NB denotes the negative binomial distribution. The mode is obtained as $\lfloor c_* (k_*-1)\rfloor$ if $k_*>1$ else 0.

\section{Proof of \sref{Corollary}{thm:pred_var}}
\label{appx:proof_thm_pred_var}
Since the site approximations of both EP and QP are Gaussian, we may analyse the predictive variances using results from the regression with Gaussian likelihood function case, namely the well known Equation (3.61) in \citep{Rasmussen:2005:GPM:1162254}:
\begin{align}
    \sigma^2 (f_*) = k(\bm x_*,\bm x_*)-\bm k_*^{\T}(K+\widetilde{\Sigma})^{-1}\bm k_*,
    \label{eq:pred_var}
\end{align}
where $f_*=f(\bm x_*)$ is the evaluation of the latent function at $\bm x_*$ and $\bm k_* = [k(\bm x_*, \bm x_1),\cdots,k(\bm x_*,\bm x_N)]^{\T}$ is the covariance vector between the test data $\bm x_*$ and the training data $\{\bm x_i\}_{i=1}^{N}$, $K$ is the prior covariance matrix and $\widetilde{\Sigma}$ is the diagonal matrix with elements of site variances $\widetilde{\sigma}_i^2$.

After updating the parameters of a site function $t_i(f_i)$, the term $(K+\widetilde{\Sigma})^{-1}$ is updated to $(K+\widetilde{\Sigma}+(\widetilde{\sigma}^2_{i,\text{new}}-\widetilde{\sigma}^2_i)\bm e_i \bm e_i^{\T})^{-1}$ where $\widetilde{\sigma}_{i,\text{new}}$ is the site variance estimated by EP or QP and $\bm e_i$ is a unit vector in direction $i$. Using the Woodbury, Sherman \& Morrison formula \citep[A.9]{Rasmussen:2005:GPM:1162254}, we rewrite $(K+\widetilde{\Sigma}+(\widetilde{\sigma}^2_{i,\text{new}}-\widetilde{\sigma}^2_i)\bm e_i \bm e_i^{\T})^{-1}$ as 
\begin{align}
    &(K+\widetilde{\Sigma}+(\widetilde{\sigma}^2_{i,\text{new}}-\widetilde{\sigma}^2_i)\bm e_i \bm e_i^{\T})^{-1} 
    \\
    &\equiv (A^{-1}+(\widetilde{\sigma}^2_{i,\text{new}}-\widetilde{\sigma}^2_i)\bm e_i \bm e_i^{\T})^{-1}
    \\
    &\quad=A-A\bm e_i [(\widetilde{\sigma}^2_{i,\text{new}}-\widetilde{\sigma}^2_i)^{-1}+\bm e_i^{\T}A\bm e_i]^{-1} \bm e_i^{\T}A
    \\
    &\quad \equiv A-\bm s_i[(\widetilde{\sigma}^2_{i,\text{new}}-\widetilde{\sigma}_i^2)^{-1}+A_{ii}]^{-1} \bm s_i^{\T}
    \\
    & \qquad = A-\frac{1}{(\widetilde{\sigma}^2_{i,\text{new}}-\widetilde{\sigma}^2_i)^{-1}+A_{ii}} \bm s_i \bm s_i^{\T}
\end{align}
where $A = (K+\widetilde{\Sigma})^{-1}$ and $\bm s_i$ is the $i$'th column of $A$. Putting the above expression into \eqnref{eq:pred_var}, we have that the predictive variance is updated according to:
\begin{align}
\sigma_{\text{new}}^2(f_*) = k(\bm x_*,\bm x_*)-\bm k_*^{\T} A \bm k_*+ \frac{1}{(\widetilde{\sigma}^2_{i,\text{new}}-\widetilde{\sigma}^2_i)^{-1}+A_{ii}}\bm k_*^{\T} \bm s_i \bm s_i^{\T}\bm k_*.
\end{align}
In EP and QP, the first two terms on the r.h.s. of the above equation are equivalent. As the site variance provided by QP is less or equal to that by EP, \ie , $\widetilde{\sigma}^2_{i,\text{QP}} \leq \widetilde{\sigma}^2_{i,\text{EP}}$, the third term on the r.h.s. for QP is less or equal to that for EP. Therefore, the predictive variance of QP is less or equal to that of EP: $\sigma_{\text{QP}}^2(f_*) \leq \sigma_{\text{EP}}^2(f_*)$.

\section{Lookup Tables}
\label{sec:lookuptables}
To speed up updating variances $\sigma_{\mathrm{QP}}^{2}$ in QP, we pre-compute the integration in \eqnref{eq:optimal_sigma} over a grid of cavity parameters $\mu$ and $\sigma$, and store the results into lookup tables. Consequently, each update step obtains $\sigma_{\mathrm{QP}}^2$ simply based on the lookup tables. Concretely, for the GP binary classification, we compute \eqnref{eq:optimal_sigma} with $\mu$, $\sigma$ and $y$ varying from -10 to 10, 0.1 to 10 and $\{-1,1\}$ respectively. $\mu$ and $\sigma$ vary in a linear scale and a log10 scale respectively, and both have a step size of 0.001. The resulting lookup tables has a size of  $20001\times 2001$. In a similar way, we make the lookup table for the Poisson regression. In the experiments, we exploit the linear interpolation to fit $\sigma_{\mathrm{QP}}^{2}$ given $\mu \in [-10,10]$ and $\sigma \in [0.1,10]$, and if $\mu$ and $\sigma$ lie out of the lookup table,  $\sigma_{\mathrm{QP}}^{2}$ is approximately computed by the EP update formula, i.e., $\sigma_{\mathrm{QP}}^{2}\approx \sigma_{\mathrm{EP}}^{2}$. On Intel(R) Xeon(R) CPU E5-2680 v4 @ 2.40GHz, we observe the running time of EP and QP is almost the same.


\begin{algorithm}[t]
    \begin{algorithmic}[1]
    \caption{Expectation (Quantile) Propagation} \label{alg:ep}
    \REQUIRE $p(\bm f)$, $p(y_i|f_i)$, $t_i(f_i)$, $i=1,\cdots,N$, $\bm \theta$
    \ENSURE $q(\bm f)$ \rtext{approximate posterior }
    \REPEAT
    \STATE compute $q(\bm f) \propto p(\bm f) \prod_i t_i(f_i)$  \rtext{by \eqref{eq:approximate_posterior}}
    \REPEAT
    \FOR{$i = 1$ to $N$}
    \STATE compute $q^{\backslash i}(f_i) \propto q(f_i)/t_i(f_i)$ \rtext{cavity}
    \STATE compute $\widetilde{q}(f_i)\propto q^{\backslash i}(f_i)p(y_i| f_i)$ \rtext{tilted}
    \IF{EP}
    \STATE $t_i(f_i) \propto \text{proj}_{\KL}[\widetilde{q}(f_i)]/q^{\backslash i}(f_i)$ \rtext{by \eqref{eq:ep_update}\eqref{eq:ep_update2}}
    \ELSIF{QP} 
    \STATE $t_i(f_i) \propto \text{proj}_{\W}[\widetilde{q}(f_i)]/q^{\backslash i}(f_i)$ \rtext{by \eqref{eq:optimal_sigma}\eqref{eq:ep_update2}}
    \ENDIF
    \STATE  update $q(\bm f) \propto p(\bm f) \prod_i t_i(f_i)$ \rtext{by \eqref{eq:approximate_posterior}}
    \ENDFOR
    \UNTIL convergence
    \STATE $\bm \theta = \argmax_{\bm \theta} \log q(\mD)$ \rtext{by \eqref{eq:log_pd}}
    \UNTIL convergence
    \STATE \textbf{return} $q(\bm f)$
    \end{algorithmic}
    \end{algorithm}